%% file: replica.tex
\documentclass[12pt]{iopart}

\usepackage[]{graphicx}
\usepackage{bm}
\usepackage{setstack}
\usepackage{iopams}
\usepackage{xifthen}
\usepackage{xargs}

\def\x{\bm{x}}
\def\y{\bm{y}}
\def\f{\bm{f}}
\def\g{\bm{g}}
\def\h{\bm{h}}
\def\y{\bm{y}}
\def\x{\bm{x}}
\def\K{\bm{K}}
\def\D{\bm{D}}
\def\C{\bm{C}}
\def\A{\bm{A}}
\def\M{\bm{M}}
\def\T{^{\rm{T}}}
\def\S{\bm{S}}
\def\e0{\bm{e}_0}
\def\X{\bm{X}}
\def\O{\bm{O}}
\def\hh{\hat{h}}
\def\bhh{\bm{\hh}}
\def\hrho{\hat{\rho}}

\def\hpvec{\bm{\tilde{h}}}
\def\hpvecvec{\bm{\underline{\tilde{h}}}}
\def\rhofunc{\rho(\hpvec^1,\ldots,\hpvec^n)}
\def\rhofuncdash{\rho( (\hpvec^1)',\ldots,(\hpvec^n)')}
\def\rholocfunc{\rho(\hpvec^1,\ldots,\hpvec^{n},\hpvec\aux^{1,1},\ldots,\hpvec\aux^{n,L},\kappa^1,
\ldots,\kappa^n,\hkappa^{1},\ldots,\hkappa^{n})}
\def\hrhofunc{\hrho(\hpvec^1,\ldots,\hpvec^n)}
\def\rhofuncshort{\rho(\ldots)}
\def\rhofuncshortdash{\rho(\ldots')}
\def\hrhofuncshort{\hrho(\ldots)}
\def\hrhofuncsiteshort{\hrho(\ldots_{i})}
\def\hrhofuncsite{\hrho(\hpvec^1_i,\ldots,\hpvec^{n}_i)}

\def\hpsi{\hat{\psi}}
\def\hpi{\hat{\pi}}
\def\hPsi{\hat{\Psi}}

\def\V{\bm{V}}
\def\hkappa{\hat{\kappa}}
\def\bhkappa{\bm{\hkappa}}
\def\bkappa{\bm{\kappa}}

\def\hPhi{\hat{\Phi}}

\def\Var{\mathop{\rm Var}} 
\def\hd{\hat{d}}
\def\bhd{\bm{\hd}}
\def\hmu{\hat{\mu}}
\def\Hmat{\bm{H}}
\def\hphi{\hat{\phi}}
\newcommand{\aux}[1][]{\ifthenelse{\isempty{#1}}{_{\mathrm{aux}}}{_{\mathrm{aux},\scriptstyle{#1}}}}
\def\auxrm{{\mathrm{aux}}}
\newcommand{\mar}[1][]{\ifthenelse{\isempty{#1}}{^{\mathrm{mar}}}{^{\mathrm{mar},\scriptstyle{#1}}}}
\newcommand{\fin}[1][]{\ifthenelse{\isempty{#1}}{_{\mathrm{fin}}}{_{\mathrm{fin},\scriptstyle{#1}}}}
\newcommand{\psifunc}[1][]{\ifthenelse{\isempty{#1}}{\psi(\hpvec)}{\psi(\hpvec^{#1})}}
\newcommand{\psifuncdash}[1][]{\ifthenelse{\isempty{#1}}{\psi'(\hpvec')}{\psi'( (\hpvec^{#1})')}}

\newcommandx*\hpsifunc[2][1,2]{
\ifthenelse{\isempty{#1}}{\ifthenelse{\isempty{#2}}{\hpsi(\hpvec)}{\hpsi^{#2}(\hpvec)}}
{\ifthenelse{\isempty{#2}}{\hpsi(\hpvec^{#1})}{\hpsi^{#2}(\hpvec^{#1})}}
}

\newcommandx*\hpsirenormfunc[2][1,2]{
\ifthenelse{\isempty{#2}}{\ifthenelse{\isempty{#1}}{\hpsi(\hpvec,\kappa,\hkappa,\{\hpvec\aux^l\})}{\hpsi(\hpvec^{#1},\kappa^{#1},\hkappa^{#1},\{\hpvec\aux^{l,#1}\})}}
{\ifthenelse{\isempty{#1}}{\hpsi^{#2}(\hpvec,\kappa,\hkappa,\{\hpvec\aux^l\})}{\hpsi^{#2}(\hpvec^{#1},\kappa^{#1},\hkappa^{#1},\{\hpvec\aux^{l,#1}\})}}
}

\newcommandx*\hphiindeprenormfunc[2][1,2]{
\ifthenelse{\isempty{#2}}{\ifthenelse{\isempty{#1}}{\hphi(\hpvec,\{\hpvec\aux^l\})}{\hphi(\hpvec^{#1},\{\hpvec\aux^{l,#1}\})}}
{\ifthenelse{\isempty{#1}}{\hphi^{#2}(\hpvec,\{\hpvec\aux^l\})}{\hphi^{#2}(\hpvec^{#1},\{\hpvec\aux^{l,#1}\})}}
}

\newcommandx*\phiindeprenormfuncdash[1][1]{
\ifthenelse{\isempty{#1}}{\phi(\hpvec',\{(\hpvec\aux^l)'\})}{\phi^{#1}(\hpvec',\{(\hpvec\aux^{l})'\})}
}

\newcommand{\psirenormfunc}[1][]{\ifthenelse{\isempty{#1}}{\psi(\hpvec,\kappa,\hkappa,\{\hpvec\aux^l\})}{\psi(\hpvec^{#1}
,\kappa^{#1},\hkappa^{#1},\{\hpvec\aux^{l,a}\})}}

\newcommand{\psirenormfuncdash}[1][]{\ifthenelse{\isempty{#1}}{\psi'(\hpvec',\kappa',\hkappa',\{(\hpvec\aux^l)'\})}{\psi'( (\hpvec^{#1})',(\kappa^{#1})',(\hkappa^{#1})',\{(\hpvec\aux^{l,#1})'\})}}

\newcommand{\phiindeprenormfunc}[1][]{\ifthenelse{\isempty{#1}}{\phi(\hpvec,\{\hpvec\aux^l\})}{\phi(\hpvec^{#1},\{\hpvec\aux^{l,a}\})}}

\newcommand{\phifuncdash}[1][]{\ifthenelse{\isempty{#1}}{\phi'(\hpvec',\{(\hpvec\aux^l)'\})}{\phi'(\hpvec^{#1},\{\hpvec\aux^{l,a}\})}}

\renewcommand{\ni}[1][]{\ifthenelse{\isempty{#1}}{\gamma}{\gamma_{#1}}}
\newcommand{\haux}[1][]{\ifthenelse{\isempty{#1}}{h_{\rm{aux}}}{h_{\rm{aux},#1}}}
\newcommand{\hhaux}[1][]{\ifthenelse{\isempty{#1}}{\hh_{\rm{aux}}}{\hh_{\rm{aux},#1}}}
\def\localfunc{\bm{W}}

\newlength{\Wspace}
\newlength{\SymWdth}

\newcommandx*{\rev}[3][2,3]{\ifthenelse{\isempty{#2}\AND\isempty{#3}}
{
\overleftarrow{#1}
}
{
\settowidth{\Wspace}{$\overleftarrow{#1}$}
\settowidth{\SymWdth}{$#1$}
\addtolength{\Wspace}{-\SymWdth}
\ifthenelse{\isempty{#3}}{
\leavevmode \rlap{\overleftarrow{\phantom{#1}}}{\hspace*{.5\Wspace}{#1}_{#2}}
}
{
\ifthenelse{\isempty{#2}}
{
\leavevmode \rlap{\overleftarrow{\phantom{#1}}}{\hspace*{.5\Wspace}{#1}^{#3}}
}
{
\leavevmode \rlap{\overleftarrow{\phantom{#1}}}{\hspace*{.5\Wspace}{#1}_{#2}^{#3}}
}
}
}}

\begin{document} 
\title[Learning curves of Gaussian process regression on random graphs]{Replica theory for learning curves for Gaussian processes on random graphs} 
\author{M J Urry and P Sollich}
\address{King's College London,
Department of Mathematics,
Strand,
London,
\mbox{WC2R 2LS},
U.K. }
\eads{\mailto{matthew.urry@kcl.ac.uk}, \mailto{peter.sollich@kcl.ac.uk}}

\begin{abstract}
Statistical physics approaches can be used to derive accurate
predictions for the performance of inference methods learning from
potentially noisy data, as quantified by the learning curve defined as the
average error versus number of training examples. We analyse a
challenging problem in the area of non-parametric inference where an
effectively infinite number of parameters has to be learned,
specifically Gaussian process regression. When the inputs are vertices
on a random graph and the outputs noisy function values, we show that replica techniques can be used to obtain exact
performance predictions in the limit of large graphs. The covariance of the
Gaussian process prior is defined by a random walk kernel, the
discrete analogue of squared exponential kernels on continuous
spaces. Conventionally this kernel is normalised only globally, so that
the prior variance can differ between vertices; as a more principled
alternative we consider local normalisation, where the prior variance
is uniform.
%
%
The starting point is to represent the average error as the derivative
of an appropriate partition function. We rewrite this in terms of a
graphical model, where only neighbour vertices are directly
coupled. Treating the average over training examples and random graphs
in a replica approach then yields learning curve predictions for the
globally normalised kernel. The results apply generically to all
random graph ensembles constrained by a fixed but arbitrary degree
distribution. For the locally normalised kernel, the normalisation
constants for the prior have to be defined as thermal averages in an
unnormalised model. This case is therefore technically more difficult
and requires the introduction of a second, auxiliary set of replicas.
We compare our predictions with numerically simulated learning curves
for two paradigmatic graph ensembles: Erd\H{o}s-R\'{e}nyi graphs with
a Poisson degree distribution, and an ensemble with a power law degree
distribution. We find excellent agreement between our predictions and
numerical simulations. We also compare our predictions to existing learning curve
approximations. These show significant deviations; we analyse briefly
where these arise.
\end{abstract}

\maketitle

\section{Introduction}

Ever since the seminal work of Seung, Sompolinsky and
Tishby~\cite{Seung1992}, it has been recognised that statistical physics can 
make significant contributions to the understanding of methods and
algorithms that learn from examples. The performance of a learning
algorithm or inference method is captured in the \emph{learning
curve}, which records the average prediction error over all training sets of a given
size, or \emph{generalisation error}, as a function of the number of training
examples.

Learning curves have been successfully analysed using statistical physics for a variety of parametric learning methods (where a finite number of parameters must be learned) by taking advantage of the interpretation of the average over training sets as quenched disorder~\cite{Seung1992,Amari1992,Watkin1993,Opper1995,Haussler1996,Freeman1997}. 
Rather more challenging is the non-parametric case, where the number of parameters is effectively
infinite. An important example in this class is the learning of functions
(regression) with \emph{Gaussian process (GP)} priors. Here the set of
parameters to be learned is in essence the entire underlying function
that one is trying to estimate from the data. GPs have been widely
adopted in the machine learning community as efficient and flexible
inference methods. This is due to a few important advantages, namely
that prior assumptions about the function to be learnt can be
seamlessly encoded in a transparent way, and that inference (at least
in the case considered in this paper) is relatively straightforward.

The general regression setting is as follows. We are given a set of
data \mbox{$(\x,\y)$} consisting of
a set of inputs $\x=(x_{1},\ldots,x_{N})\T$ and a corresponding vector of outputs $\y=(y_1,\ldots,y_{N})\T$ and wish to infer the underlying
true function $y=f(x)$. In a Bayesian approach we infer not a single
function $f$, but a posterior distribution, $P(f|\x,\y)$, over a function
space. The posterior distribution is calculated from two elements: a
likelihood $P(\y|f,\x)$ which measures the probability the outputs $\y$ were
generated by a particular function $f$, and a prior distribution $P(f)$
which encodes any prior beliefs about how plausible different
functions would be in the absence of data.  The prior can encode
properties such as the expected smoothness of the function, its
typical amplitude, or the lengthscale on which it varies across the
input space. The posterior is calculated from the prior and likelihood
using Bayes' theorem:
\begin{equation}
P(f|\x,\y) = \frac{P(\y|f,\x)P(f)}{\int \rmd f' P(\y|f',\x)P(f')}.
\end{equation}
For the particular case of GPs, the prior is assumed to be a Gaussian
process~\cite{Rasmussen2005}. This means that any set of function values
$f(x_1),\ldots,f(x_k)$ is assumed to have a joint Gaussian
distribution. The statistics of this Gaussian distribution, and hence
the Gaussian process, are specified by a \emph{covariance function} or
\emph{kernel} $C(x,x')$ and a \emph{mean function}, $\mu(x)$.  The
kernel gives the covariance $\langle f(x)f(x')\rangle$ under the prior
between the values of $f$ at input points $x$ and $x'$. It is the
kernel that gives GPs and other kernel based methods their intuitive
nature as it describes in a simple manner prior assumptions such as
smoothness, lengthscale of variation and typical amplitude of the
function we aim to learn. The mean function $\mu(x)$ gives the average
of $f(x)$ under the prior, and is normally set to zero unless one has
very strong prior knowledge about a non-zero function mean.

We will consider the simplest case for the likelihood: that data
points $y_\mu$ are generated by corrupting 
$f(x_{\mu})$ with normally distributed noise, i.e.\
\mbox{$P(\y|f,\x)\propto\exp(-\sum_{\mu} 
(f(x_\mu) - y_{\mu})^{2})/(2\sigma^2)$}. Then the posterior distribution is again
a Gaussian process, and the posterior mean and variance of $f(x)$ can
be expressed in closed form as~\cite{Rasmussen2005}
\begin{eqnarray}
  \bar{f}(x^{*}) =\bm{k}(x^{*})\T\bm{\mathcal{K}}^{-1}\bm{y}\\ 
  \Var(f(x^{*})) =
C(x^{*},x^{*})-\bm{k}(x^{*})\T\bm{\mathcal{K}}^{-1}\bm{k}(x^{*}),
\label{eqn:GPvariance}
\end{eqnarray}
with the matrix $\bm{\mathcal{K}}$ and vector $\bm{k}$ having entries
$\mathcal{K}_{ij} = C(x_i,x_j) + \delta_{ij}\sigma^{2}$ and 
$k_{i}(x^{*}) = C(x_{i},x^{*})$ respectively.
One would then use $\bar{f}(x^*)$ as the prediction for the function
value at a test input $x^*$, and the square root of the posterior
variance can be used to give an error bar for this prediction.

Analysing the average performance of GPs at predicting a function from
data, \mbox{i.e.\ calculating} the learning curve, is an interesting problem
that has been well studied for the case of functions defined over
continuous spaces~\cite{Sollich1999a, Sollich1999b,
Opper1999, Williams2000, Malzahn2003, Sollich2002a,
Sollich2002b, Sollich2005}. However, far less is understood about the
learning curves for GP regression for functions defined on discrete
spaces. With the rise of large structured data such as the internet,
social networks, protein networks and financial markets,
such an understanding of the performance of GPs, and machine learning
techniques in general, on discrete spaces is becoming important.

In this paper we concern ourselves with obtaining predictions for the learning curves of GP regression for functions on large random graphs. We study GPs trying to predict a
function $f:\mathcal{V}\to\mathbb{R}$ defined on the vertices of a
graph $G(\mathcal{V},\mathcal{E})$ with vertex set $\mathcal{V}$ and
edge set $\mathcal{E}$. As is standard in the analysis of GP learning
curves we will
focus on GPs with zero mean. (Our results can easily be
generalised to non-zero mean, but at the expense of increased notational
complexity which we avoid here.) We note that, in the context of GPs
on graphs, the covariance function will be a $V\times V$
covariance matrix where $V=|\mathcal{V}|$ is the number of vertices of
the graph.

We focus on GPs with covariance structure described by a \emph{random
walk kernel} first introduced in \cite{Kondor2002} and further
developed in \cite{Smola2003}. These are generalisations of the
frequently used square exponential kernels in continuous spaces (see
for example \cite{Rasmussen2005}). We define the random walk kernel
by:
\begin{equation}\label{eqn:rw}
  \eqalign{
  \C &= \K^{-1/2}\left(\bm{I} - a\bm{L}\right)^{p}\K^{-1/2}\\
  &= \K^{-1/2}\left( (1-a^{-1})\bm{I} +
  a^{-1}\D^{-1/2}\A\D^{-1/2}\right)^{p}\K^{-1/2},}
\end{equation}
where $a$ and $p$ are hyperparameters of the kernel, $\K =
\rm{diag}(\kappa_1,\dots\kappa_V)$ is a normaliser which controls the
prior variance of $f$, $\bm{I}$ is the identity matrix, $\A$ the
adjacency matrix of $G$ ($a_{ij}=1$ if $(i,j)\in \mathcal{E}$,
$a_{ij}=0$ otherwise), $\D$ is the diagonal degree matrix with $D_{ii} =
\sum_{j}a_{ij}$ and $\bm{L} = \bm{I} - \D^{-1/2}\A\D^{-1/2}$ the
normalised Laplacian \cite{Chung1996}. 
(The Laplacian defined in \cite{Chung1996} differs slightly, in that
for single vertices $i$ that are disconnected from the rest of the
graph it has $L_{ii}=0$. Our definition has $L_{ii}=1$ for this case. In the case
of graphs without single disconnected vertices the two definitions of the normalised Laplacian will agree. We use this adjusted normalised Laplacian because analysis is easier without the special case and for local normalisation of the kernel, for which we will shortly argue in favour of, the resulting normalised covariance function will be the same.)
\Eref{eqn:rw} is called the random walk kernel because it defines
the importance of neighbouring vertices in predicting a function value
at some given vertex $i$ in terms of the frequency with which a lazy
random walk starting from $i$ lands at the neighbour. We can view
$a^{-1}$ as the probability of making a `step' and $p$ the number of
attempts the walker makes. With this interpretation $p/a$, the average
number of steps of the lazy random walk, can be thought of as the
\emph{lengthscale} of \eref{eqn:rw}
\cite{Sollich2009}.

We will consider two normalisations of the random walk kernel:
\emph{global normalisation} and \emph{local normalisation}.
In the case of global normalisation we fix all $\kappa_i$ to be equal,
chosen such that the average prior variance $(1/N)\sum_{i=1}^V C_{ii}$
has the desired value which we fix to unity. For local
normalisation, we set the $\kappa_i$ so that the function value at each
vertex has the same prior variance, namely $C_{ii}=1$. It is worth noting, that
from a Bayesian modelling point of view, local normalisation is more
natural since we do not expect to have strong prior knowledge that
would justify using different prior variances at different
vertices. However, we also analyse the case of global normalisation
because that is the form for the kernel suggested in
\cite{Kondor2002,Smola2003} and used in
e.g.~\cite{Min2009,Lippert2010,Gao2009}. This case also 
serves to illustrate the general structure of the replica approach we
use for the analysis, before we tackle the rather more challenging scenario of local
normalisation.

Our main aim in this paper is to show that for the setting of GP
regression of functions on graphs we can calculate the learning curve
exactly in the limit of large graphs. Our results will be valid for
graphs chosen from a broad range of random graph ensembles, which are
defined by an arbitrary degree distribution. This is rather remarkable: in
the case of GP regression of functions defined on continuous spaces,
exact learning curve predictions are possible only in very special
cases~\cite{Malzahn2005,Rasmussen2005}. Otherwise, one has to rely on      
approximations that typically give only qualitatively accurate
predictions, and can fail dramatically in some scenarios, e.g.\ for low
noise level $\sigma^2$.

The rest of this paper is structured as follows: we begin in
\sref{sec:partfunction} by defining the generalisation error and
expressing it in terms of a generating partition function. This is
then rewritten in a form that can act as a suitable starting point for
a replica approach, using additional variables to obtain a model with
interactions only between neighbouring vertices. We briefly compare
our method to a similar approach by Opper~\cite{Opper2002} and point
out when our method and that of Opper's diverge. In
\sref{sec:repanalysis} we use a replica approach to calculate the
learning curves. We show in
\sref{sec:global} that the case of global normalisation can be solved
by applying a method similar to that used in~\cite{Kuhn2007}.  In
\sref{sec:local} we analyse learning curves for the
locally normalised kernel. We show that the introduction of local normalisation
brings a great deal more
complexity to the problem: we need to use an additional set of
replicas to account for the fact that the normalisation is now tied
tightly but in a non-local way to the quenched disorder given by the
graph structure. The statistics of the auxiliary replicas are obtained
by a saddle point approach that is nested inside the conventional
saddle point method for the ordinary replicas.
\Sref{sec:results} compares our replica predictions to
numerically simulated learning curves and existing learning curve
approximations~\cite{Sollich1999a,Opper2002}, suitably extended from
continuous to discrete input spaces. We also discuss the qualitative
differences between the learning curves produced by the globally and
locally normalised kernels. Finally
\sref{sec:conc} summarises our results and discusses
potential avenues for future research.

\section{Learning curves} \label{sec:partfunction}

To begin with we express the generalisation error for GP regression on
a large random graph as the derivative of an appropriate partition
function $Z$. We will introduce additional variables to get this into
the form of an \emph{undirected graphical
  model}. By this we
  mean, following the terminology in the machine learning literature,
  a Boltzmann distribution with a Hamiltonian containing, in addition
to local `energy' contributions for each vertex, only pairwise
interaction energies between neighbouring vertices on the graph.
By making some assumptions on the graph structure, namely that the
graph is random subject to an arbitrary specified degree distribution,
we will be able to apply a replica method to calculate the dependence
of the generalisation error on the number of training examples, i.e.\
the learning curve. Physically, the reason that this procedure can be
carried through is the fact that random graphs from the ensembles
considered here typically have locally tree-like
structures. Accordingly, the results that we find can also be derived
using a cavity method~\cite{UrryTBA,Urry2010}.

The generalisation error of a GP is defined as the squared deviation
between the predicted function, i.e.\ the student's posterior mean,
$\langle\f\rangle_{\f|\x,\y}$, and 
the true teacher function $\g$ that generated the data. This is then
averaged over data sets, i.e.\ noise corrupted outputs $y_\mu$
generated from the true function $\g$ for a fixed set of inputs $\x$,
and independently and identically distributed inputs $x_\mu$.
Finally one averages over the prior distribution of teachers $\g$
(in this paper assumed to be a GP) and,
in our random graph setting, the graph ensemble $\mathcal{G}$:
\begin{equation}\label{eqn:epgdef}
    \epsilon_{g} = \Bigg\langle \Bigg\langle\Bigg\langle\Bigg\langle\frac{1}{V}\sum_{i=1}^{V}\left(g_i - \langle
f_i\rangle_{\f|\x,\y}\right)^{2}
    \Bigg\rangle_{\y|\g,\x}\Bigg\rangle_{\x}\Bigg\rangle_{\g}\Bigg\rangle_{\mathcal{G}}.
\end{equation}
Here we have defined $\f$ and $\g$ to be $V$-dimensional vectors with $f_{i} = f(x_i)$ and $g_{i} 
= g(x_i)$.

In this paper we will assume, as is typical (though see \cite{Malzahn2005}) in learning curve  
calculations, that teacher and student have the same posterior
distribution, i.e.\ a  
\emph{matched scenario}. The generalisation error in this case is also
the Bayes error, i.e.\ the lowest average error achievable for a
teacher function from the given GP prior. Under this assumption
\eref{eqn:epgdef} simplifies to just the posterior variance of the
student (see for example \cite{Rasmussen2005}). Since we require only
the posterior variance we shift our functions so that $f_i$ now represents
the deviation of the function $\f$  from the posterior
mean. The generalisation error can then be expressed as
\begin{equation}\label{eqn:epgshift}
  \epsilon_{g} = \Bigg\langle\Bigg\langle\Bigg\langle
\frac{1}{V}\sum_{i=1}^{V}f_i^{2}\Bigg\rangle_{\f|\x}\Bigg\rangle_{\x}\Bigg\rangle_{\mathcal{G}}.
\end{equation}
The posterior for
  the redefined $\f$ is proportional to the product of the prior
$P(\f)$ and the quadratic exponential terms from the likelihood
$P(\y|\f,\x)$. The latter are $\exp(-\sum_{\mu} 
f_\mu^{2}/(2\sigma^{2}))=\exp(-\sum_{i=1}^{V}\ni[i]f_{i}^2/(2\sigma^{2}))$ where 
$\ni[i]$ counts the number of examples seen at vertex $i$. Note that
in line with the expression \eref{eqn:GPvariance} for the GP variance not  
depending on the training outputs, the posterior distribution for the shifted
function values only depends on the training inputs, hence the notation $f|\x$ rather than $f|\x,\y$ in
\eref{eqn:epgshift}. For the same reason we have also been able to
drop the averages over $\y$ and $\g$.

We now wish to rewrite \eref{eqn:epgshift} in terms of a graphical
model. Taking a statistical physics approach we rewrite the
generalisation error in terms of a generating partition function $Z$:
\begin{eqnarray}
  \epsilon_{g} = -\frac{2}{V}\lim_{\lambda\to0}\frac{\partial}{\partial\lambda}\langle\log Z \rangle_{\x,
  \mathcal{G}}\label{eqn:epg_Z}\\
  Z = \int \rmd\f \exp\left(-\frac{1}{2}\f\T\C^{-1}\f-\frac{1}{2\sigma^{2}}\sum_{i=1}^{V}\ni[i]f_{i}^2 - 
  \frac{\lambda}{2}\f\T\f\right).\label{eqn:Zdef}
\end{eqnarray}
To bring this into a graphical model form we need to
untangle the relationship between function values at different
vertices arising from the prior. We first rewrite the prior term as an
integral over its Fourier transform in order to eliminate the inverse
of the covariance function. Carrying out the remaining, now factorised,
integral over the $f_i$ we are left with
\begin{equation}\label{eqn:Zft}
  Z = |\S|^{1/2}\int \rmd\h \exp\left(-\frac{1}{2}\h\T\C\h-\frac{1}{2}\h\T\S\h\right),
\end{equation}
where $\S = \rm{diag}\left(1/(\frac{\ni[1]}{\sigma^{2}} +
\lambda),\dots,1/(\frac{\ni[V]}{\sigma^{2}} + \lambda)\right)$ and we
have dropped constant factors that do not depend on $\lambda$. If
$p=1$ we would now have the required graphical model form.
For larger $p$, however, the first term in
\eref{eqn:Zft} still relates more than nearest 
neighbour vertices. We can make headway in dealing with the complicated interactions introduced by $\C$ 
by rewriting \eref{eqn:rw} in terms of a binomial expansion,
\begin{equation}
    \C = \sum_{q=0}^{p}{p \choose q}(a^{-1})^{q}(1-a^{-1})^{p-
    q}\K^{-1/2}\left(\D^{-1/2}\A\D^{-1/2}\right)^{q}\K^{-1/2}.
\end{equation}
We can now introduce $\h^{q} =
\left(\K^{1/2}\D^{-1/2}\A\D^{-1/2}\K^{-1/2}\right)^{q}\h$ into
\eref{eqn:Zft} recursively, by setting $\h^{0}=\h$ and enforcing 
$\h^{q} =
\K^{1/2}\D^{-1/2}\A\D^{-1/2}\K^{-1/2}\h^{q-1}$ for $q=1,\ldots,p$
using
a Fourier transformed delta function. Our partition function then becomes
\begin{equation}\label{eqn:Zfinal}
  \eqalign{
  \fl
  Z = |\S|^{1/2}\int \prod_{q=0}^{p}\rmd\h^q \prod_{q=1}^{p}\rmd\bhh^q \exp\left(-\frac{1}
  {2}\sum_{q=0}^{p}c_q(\h^{0})\T\K^{-1}\h^{q}
  -\frac{1}{2}(\h^{0})\T\S\h^{0}\right.\\
  \left.+\rmi\sum_{q=1}^{p}(\bhh^q)\T\h^q- \rmi\sum_{q=1}^{p}
  (\bhh^{q})\T\K^{1/2}\D^{-1/2}\A\D^{-1/2}\K^{-1/2}\h^{q-1}\right),}
\end{equation}
with $c_{q} = {p \choose q}(a^{-1})^{q}(1-a^{-1})^{p-q}$ and where we
have again dropped $\lambda$-independent factors. We now have the
desired form for $Z$. Rescaling variables in \eref{eqn:Zfinal} to $(h^*)^q_i =
h^q_id_i^{-1/2}\kappa_{i}^{-1/2}$ and $(\hh^*)^q_{i} =
\hh^q_id_i^{-1/2}\kappa_i^{1/2}$ and rewriting explicitly in terms of
vertex and interaction terms we have, after dropping the asterisks
again, 
\begin{equation}\label{eqn:Zsites}
  Z = \int \rmd\hpvecvec \prod_{i\in\mathcal{V}}\exp\left(-\mathcal{H}(\hpvec_i,d_i,\kappa_i,
  \ni[i])\right)
  \prod_{(i,j)\in\mathcal{E}}\exp\left[-\mathcal{J}(\hpvec_i,\hpvec_j)\right],
\end{equation}
with
\begin{equation}\label{eqn:singlesite}
  \fl
  \mathcal{H}(\hpvec,d,\kappa,\ni) = \frac{d}{2}\sum_{q=0}^{p}c_qh^{0}h^{q}
  +\frac{\kappa d (h^{0})^2}{2(\ni/\sigma^{2}+\lambda)}-\rmi d\sum_{q=1}^{p}
  \hh^q h^q-\frac{1}{2}\log\left(\frac{\kappa}{\ni/\sigma^{2}+\lambda}\right)
  \end{equation}
  \begin{equation}\label{eqn:siteinteraction}
    \mathcal{J}(\hpvec,\hpvec') = \rmi\sum_{q=1}^{p}\left( \hh^{q} (h')^{q-1} + ( 
    \hh')^{q}h^{q-1}\right),
  \end{equation}
$\hpvec=(h^{0},\ldots,h^{p},\hh^{1},\ldots,\hh^{p})\T$ and $\hpvecvec=(\hpvec_{1},\ldots,\hpvec_V)$. \Eref{eqn:Zsites} is now in the form of a (complex-valued) Gaussian
graphical model on a graph with 
local fields and interactions only between nearest neighbour pairs.

We can now proceed in the standard way using replicas. Since
$\langle\log Z \rangle_{\x,\mathcal{G}}$ is hard to calculate we use
the replica trick, $\langle\log Z \rangle_{\x,\mathcal{G}} = \lim_{n\to
0}\frac{1}{n}\log\langle Z^{n}\rangle_{\x,\mathcal{G}}$, to bring the
averages inside the logarithm. We calculate $\langle
Z^{n}\rangle_{\x,\mathcal{G}}$ for integer $n$ and perform an analytic
continuation for $n\to0$ at the end.

Even after introducing replicas, the average of the replicated
partition function over inputs (and, here, graphs) may be analytically
intractable, depending on what stage in the calculation averages are
carried out. It is here that our method and earlier approximations in
continuous spaces \cite{Sollich1999a,Malzahn2003} differ: in the latter
approaches, because the graph structure could not be exploited, the
average over inputs $\x$ has to be kept in general form. To proceed,
a Gaussian variational approximation was used in~\cite{Malzahn2003}.
One can show~\cite{UrryTBA} that for the case of GPs defined on graphs
as considered here, this variational approximation of the input
average in terms of two moments only ignores some of
fluctuations in the number of examples $\ni[i]$ seen at each
vertex. In this paper we will take a different approach. We do not attempt
to carry out the input average from the start; instead we perform
the graph average and decouple vertices by introducing replica
densities. The input average can then be kept without approximation
right until the final saddle point equations. That this is possible is
based on the following fact: if $N$ training inputs $i_\mu$ are chosen
randomly from, say, a uniform distribution over vertices, then for a
large graph ($V\to\infty$) this is equivalent to each $\ni[i]$ being
independently sampled from a Poisson distribution with mean $\nu$
where $\nu=N/V$. We can therefore carry out the input average
independently over the different vertices, without approximating the
distribution of the $\ni[i]$. This will result in predictions for the
learning curve that are exact across the entire range from small to
large $\nu$. For simplicity we restrict ourselves here to the case of
a uniform input distribution, though all results generalise
straightforwardly to the case where the input distribution is
$\omega_i/V$, with the $\omega_i$ weights of order unity that sum to $V$.

Finally in order to perform the average over the graph ensemble, we consider ensembles
consisting of all graphs with a fixed but arbitrary degree
distribution $p(d)$. To ensure that the graphs are sparsely connected
and therefore locally tree-like, we assume that the mean degree
$\bar{d}$ is finite, i.e.\ does not grow with $V$. With these
assumptions we are now able to apply the replica method to
\eref{eqn:Zsites}.

\section{Replica analysis}\label{sec:repanalysis}

Our method for calculating \eref{eqn:epg_Z} from the average of $\log
Z$ as defined in \eref{eqn:Zsites} for $V\to\infty$ follows the
approach of~\cite{Kuhn2007}. We begin by briefly considering the graph
average for the class of ensembles of graphs assumed in this paper.

For a fixed degree sequence $\{d_1,\ldots,d_V\}$, a uniform distribution over all 
graphs $G$ with the correct degree sequence is given by
\begin{equation}\label{eqn:probgraph}
  \eqalign{
  P(G|\{d_1\ldots d_V\}) = \frac{1}{\mathcal{N}}\prod_{(i,j)}p(a_{ij})\delta_{a_{ij},a_{ji}}
 \prod_{i}\delta_{d_i,\sum_{j\neq i}a_{ij}}\\
  p(a_{ij}) =\frac{\bar{d}}{V}\delta_{a_{ij},1} + \left(1-\frac{\bar{d}}{V}\right)\delta_{a_{ij},0},}
\end{equation}
with $\mathcal{N}$ defined to be the normaliser of $P(G|\{d_1\ldots
d_V\})$. All the results that we are interested in are invariant under
a relabelling of the graph vertices, and will therefore depend only on
the degree distribution  $p(d)=\frac{1}{V}\sum_i\delta_{d_i,d}$. The
mean degree is $\bar{d}=\frac{1}{V}\sum_i d_i = \sum_d p(d)d$. The
factors $p(a_{ij})$ above do not affect $P(G|\{d_1\ldots d_V\})$ as the
number of non-zero $a_{ij}$ is fixed by the degree constraints, but
simplify the calculation.
%

We split our replica analysis of the learning curve into two sections
corresponding to two different ways of normalising the covariance
kernel, as specified by the normalisation coefficients $\kappa_i$. In
\sref{sec:global}, we consider a
\emph{globally normalised kernel} where $\kappa_i=\kappa$, a constant
equal to the average prior 
variance of the unnormalised kernel. This fixes the average of the
prior variances at all vertices to unity. In \sref{sec:local}, we
consider a \emph{locally normalised kernel}, 
where $\kappa_i$ is set equal to the local prior variance of vertex
$i$ of the unnormalised kernel: this forces all local prior variances
to be exactly (rather than just on average) equal to unity.

\subsection{Global normalisation}\label{sec:global}
For a large ($V\to\infty$) graph, the normalisation by the average
prior variance of the unnormalised kernel does not depend on the 
specific graph sampled. Thus we may solve \eref{eqn:singlesite} with
an arbitrary constant $\kappa_i=\kappa$. In order to calculate the
learning curve all that is then required is to set initially
$\kappa=1$, calculate the generalisation error
at $\nu=0$ and set $\kappa$ equal to the result. The generalisation
error from the globally normalised kernel can then be calculated with this
$\kappa$ for any required value of $\nu$.

The learning curve calculation for global
normalisation will follow essentially the method of \cite{Kuhn2007},
so we will keep explanations brief. Our derivation begins by raising
\eref{eqn:Zsites} to the power $n$, substituting $\kappa_i = \kappa$
and including the graph and input averages
\begin{equation}\label{eqn:Zglobalsites}
  \eqalign{
  \fl
  \langle Z^{n}\rangle_{\x,\mathcal{G}} = \left\langle\int \prod_{a=1}^{n}\rmd\hpvecvec^{a}
  \prod_{i\in\mathcal{V}}\exp\left(-\sum_{a=1}^{n}\mathcal{H}(\hpvec^a_i,d_i,\kappa,
  \ni[i])\right)\right.\\\left.\times\prod_{(i,j)\in\mathcal{E}}\exp\left(-\sum_{a=1}^{n}\mathcal{J}(\hpvec^a_i,
  \hpvec^a_j)\right)\right\rangle_{\{\ni[i]\},\mathcal{G}}.}
\end{equation}
We aim to rewrite \eref{eqn:Zglobalsites} in terms of `replica
densities' so that we may  calculate the graph average and then
decouple different vertices. We define these replica densities as
\begin{equation}\label{eqn:globaldensities}
  \rhofunc =
  \frac{1}{V}\sum_i\prod_{a=1}^{n}\delta(\hpvec^a-\hpvec^a_i)\rme^{\rmi\hd_i},
\end{equation}
and incorporate them into \eref{eqn:Zglobalsites} with a functional delta using conjugate densities 
$\hrho$. 

Substituting \eref{eqn:globaldensities} and the functional delta enforcement term into 
\eref{eqn:Zglobalsites}, performing the graph average and rearranging terms (see 
\ref{app:globalpresaddle}, where $\hd_i$ is also defined) we obtain
\begin{equation}\label{eqn:globalZpresaddle}
 \fl \langle Z^{n}\rangle_{\x,\mathcal{G}} =\frac{1}{\mathcal{N}} \int \mathcal{D}\rho \mathcal{D}\hrho 
  \exp\left[V\left(\frac{\bar{d}}{2}(S_{1}[\rho]-1) -\rmi S_{2}[\rho,\hrho] +
  S_{3}[\hrho]\right)\right],
\end{equation}
with
\numparts
\begin{eqnarray}
 \fl S_1[\rho] = \int \prod_{a=1}^{n}\rmd\hpvec^a 
 \rmd(\hpvec^a)'\rhofunc\rhofuncdash\exp\left[-\sum_{a=1}^{n}\mathcal{J}(\hpvec^a,
 (\hpvec^a)')\right]\label{eqn:globalS1}\\
 \fl S_{2}[\rho,\hrho] = \int\prod_{a=1}^{n}\rmd\hpvec^a \rhofunc\hrhofunc\label{eqn:globalS2}\\
 \fl S_{3}[\hrho] = 
 \sum_{d}p(d)\log\left\langle\int\prod_{a=1}^{n}\rmd\hpvec^a\exp\left(-\sum_{a=1}^{n}\mathcal{H}(\hpvec^a,d,
 \kappa,\ni)\right)
 \frac{\left(\rmi\hrhofunc\right)^d}{d!}\right\rangle_{\ni}.
\label{eqn:globalS3}
\end{eqnarray}
\endnumparts

We see that for $V\to\infty$, \eref{eqn:globalZpresaddle} will be
dominated by its saddle point. To proceed we must make a specific assumption
about the form of the replica densities at this saddle point. As in
\cite{Kuhn2007} we assume symmetric replicas, making the ansatz
\begin{eqnarray}
    \rhofunc = \int \mathcal{D}\psi\,\pi[\psi]\prod_{a=1}^{n}\frac{\exp\left(-\psifunc[a]\right)}
    {Z[\psi]}\label{eqn:globalansatzrho}\\
    \hrhofunc = -\rmi\bar{d}\int
    \mathcal{D}\hpsi\,\hpi[\hpsi]\prod_{a=1}^{n}\frac{\exp\left(-\hpsifunc[a][]\right)}{Z[\hpsi]},
    \label{eqn:globalansatzhrho}
\end{eqnarray}
with the convention that
\begin{equation}
Z[f] = \int\rmd \hpvec \exp\left(-f(\hpvec)\right).
\end{equation}
This amounts to assuming each single replica function has a Gibbsian
form. The average over $\pi$ and $\hpi$ can be interpreted as
representing, physically, an average over vertices $i$. 
We have included the
prefactor $-\rmi\bar{d}$ in \eref{eqn:globalansatzhrho} because this
will ensure, in the end, that both
$\pi$ and $\hpi$ are normalised probability density
functions.

We now substitute the ansatz \eref{eqn:globalansatzrho} and
\eref{eqn:globalansatzhrho} into \eref{eqn:globalS1} to
\eref{eqn:globalS3}, and expand in the number of replicas $n$ up to $O(n)$.
Terms involving $\lambda$ first feature in the $O(n)$ term of
$S_{3}[\hrho]$. Since the generalisation error \eref{eqn:epg_Z} requires a
derivative with respect to $\lambda$ we must consider both the
leading $O(n^0)$ and subleading $O(n)$ terms in calculating our saddle
point contributions. We find that the leading $O(n^0)$ terms cancel
with the graph normaliser, $\mathcal{N}$, and constrain $\pi$ and
$\hpi$ to be normalised distributions. The subleading saddle point
equations together with these $O(n^{0})$ normalisation constraints then give the
self-consistency equation (see \ref{app:globalupdate}) 
\begin{equation}\label{eqn:globalupdate}
  \pi[\psi] = \sum_{d}\frac{p(d)d}{\bar{d}}\int 
  \prod_{i=1}^{d-1}\mathcal{D}\psi^{i}\pi[\psi^{i}]\left\langle\delta(\psi -
  \Psi[\hPsi[\psi^1],\ldots,\hPsi[\psi^{d-1}]])\right\rangle_{\gamma},
\end{equation}
with
\begin{eqnarray}
  \Psi[\hpsi^{1},\ldots,\hpsi^{d-1}](\hpvec) = \sum_{i=1}^{d-1}\hpsifunc[][i] + \mathcal{H}(\hpvec,d,
  \kappa,\ni)\label{eqn:globalPsi} \\
  \hPsi[\psi'](\hpvec) = -\log\int \rmd\hpvec' \exp\left(-\psi'(\hpvec')-\mathcal{J}(\hpvec,
  \hpvec')\right).\label{eqn:globalhPsi}
\end{eqnarray}
Looking at the explicit form of $\mathcal{J}$ and $\mathcal{H}$, one
sees that these equations are solved by `energy functions'
$\psi(\hpvec)$ that contain only quadratic terms in
$\hpvec$. This of course makes sense as we started from the partition function
of a (complex) Gaussian graphical model.
We write these functions in terms of their
(complex valued) covariance matrices $\V$ so that
$\psi(\hpvec) = \frac{1}{2}\hpvec\T\V^{-1}\hpvec$.
In terms of the distribution of the $\V$, the self-consistency
equation \eref{eqn:globalupdate} becomes
\begin{equation}\label{eqn:globalvarianceupdate}
  \fl  \pi[\V] = \sum_{d}\frac{p(d)d}{\bar{d}} \int \prod_{i=1}^{d-1}\rmd\V^{i}\pi[\V^{i}]\left\langle\delta\Big(\V - 
    \Big[\O - \sum_{i=1}^{d-1}\X\V^{i}\X\Big]^{-1}\Big)\right\rangle_{\gamma},
\end{equation}
with
\begin{equation}\label{eqn:globalOX}
\setlength{\arraycolsep}{1mm}
 \fl \bm{O} = d\left(\begin{array}{cccc|ccc}
c_0 \!+\!\frac{\kappa}{\gamma/\sigma^{2} +\lambda} &
\frac{1}{2}c_1 & \dots & \frac{1}{2}c_{p} &
0 & \dots & 0 \\
\frac{1}{2}c_{1}& & & &
-\rmi & & \\
\vdots & & & &
 & \ddots & \\
\frac{1}{2}c_{p}& & & &
 & & -\rmi\\[0.5mm]
\hline
0 & -\rmi & & &
 & & \\
\vdots & & \ddots & &
 & \bm{0}_{p,p} & \\
0 & & & -\rmi & 
 & &
\end{array}\right),\, 
\bm{X} = \left(\begin{array}{cccc|ccc}
 & & & & \rmi & & \\
 & \multicolumn{2}{c}{\bm{0}_{p+1,p+1}} & & & \ddots & \\  
 & & & & & & \rmi\\
 & & & & 0 & \dots & 0\\
\hline
\rmi & & & 0 & & & \\
  & \ddots & & \vdots & & \bm{0}_{p,p}\\
 & & \rmi & 0 & & &
\end{array}\right).
\end{equation}

At first sight \eref{eqn:globalvarianceupdate} appears to have a
singularity when $\ni=0$ and $\lambda\to 0$. Using Woodbury's identity (see
\ref{app:globalwoodburyderivations}) we can eliminate this
apparent divergence and solve the self-consistency equation numerically using population
dynamics \cite{Mezard2001}. This is an iterative technique where one creates a population of 
covariance matrices and for each iteration updates a random element of
the population according to the delta function in
\eref{eqn:globalvarianceupdate}. The update is calculated 
by sampling from the degree distribution $p(d)$ of local degrees, the Poisson distribution of the local number of
examples $\nu$ and from the distribution $\pi[\V^i]$ of `incoming'
covariance matrices, the latter being approximated
by uniform sampling from the current population.

Once a numerical solution for $\pi[\V]$
has been found, the
generalisation error can be calculated from  
\begin{equation}\label{eqn:globallearningcurve} 
  \fl
  \epsilon_g(\nu) = -\lim_{\lambda\to0}\frac{\partial S^{O(n)}_{3}}{\partial\lambda} =
  \left\langle
\sum_{d}p(d)
\int \prod_{k=1}^{d} \rmd\bm{V}_k\, \pi[\bm{V}_k]\
\frac{1}{\gamma/\sigma^2 + d\kappa\e0\T\M_{d}^{-1}\e0}\right\rangle_\gamma,
\end{equation}
where we have defined $S_{3}^{O(n)}$ as the $O(n)$ terms in
the small-$n$ expansion of $S_{3}$ (see \ref{app:globalupdate}, also for
the definition of $\M_d$), defined $\e0 = (1,0,\ldots,0)\T$ and again taken  
advantage of the Woodbury identity.

\Eref{eqn:globallearningcurve} has a simple interpretation: 
once information from the rest of the graph has been folded in, any
graph vertex has an `effective prior variance' with precision 
$d\kappa(\M_d^{-1})_{00}$. This is then combined with the $\ni$ local
examples to arrive at the final posterior variance at this vertex.

\begin{figure}
\input{population.tex}
\caption{(a) Normalised histograms of the top left entries $V_{00}$ from a
    population of covariance matrices that solves the self-consistency
    condition \eref{eqn:globalvarianceupdate}, for random regular
    graphs with degree $d=3$ and $\nu=0.01$. (b) As (a) but for $\nu=1$ and
$\nu=10$, the latter just visible as a narrow peak at the left edge of
the distribution for $\nu=1$. (c) Analogue of figures (a) and (b) combined for
Erd\H{o}s-R\'enyi random graphs with average connectivity $3$.\label{fig:pop}}
\end{figure}
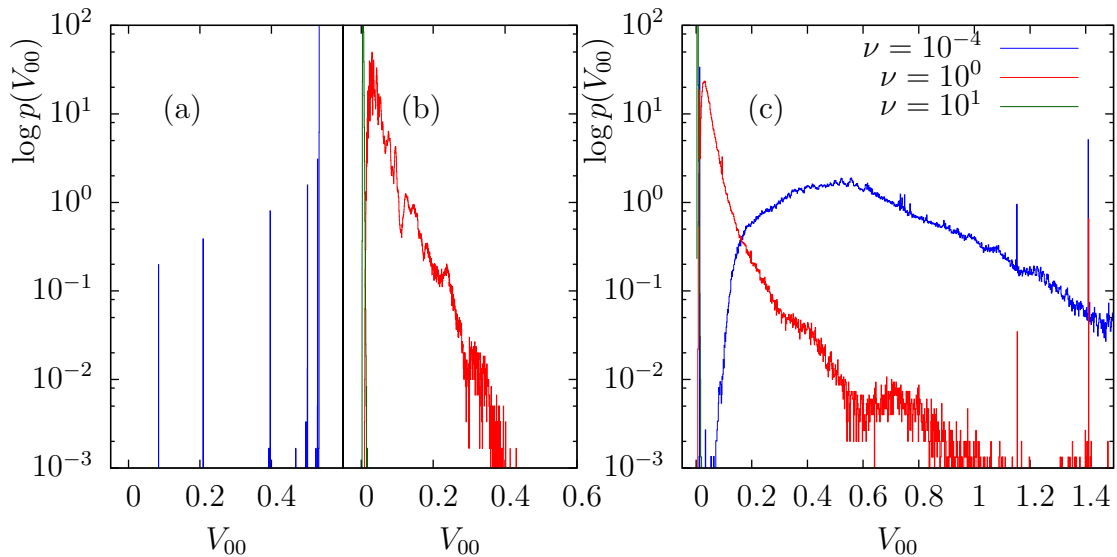

It is interesting to ask what the distribution of cavity covariances $\pi[\V]$ will be for different $\nu$. \Fref{fig:pop} 
(a) \& (b) show the log-distribution of $V_{00}$ for random regular graphs (where each vertex has fixed equal degree) with degree $d=3$.
For $\nu=0.0001$ the distribution appears to consist of delta peaks. 
This can be confirmed analytically by a Taylor expansion of
\eref{eqn:globalvarianceupdate} about $\nu=0$. Each peak can be shown to be related to the variance
of a vertex in a regular tree in which an example has been seen at
increasing distances from the vertex. One would expect the peak with
the highest amplitude
to correspond to the variance of a vertex with an example seen at a
distance $p$, because the number of vertices affected by an example grows
with distance from example vertex. Further peaks are produced
by examples at distances smaller than $p$, with height decreasing as
distance and therefore number of affected vertices
decreases. The $V_{00}$ values also decrease, since the nearer to an
example a vertex is, the more the local posterior variance is reduced.
For larger $\nu$ more and more
peaks are added until the distribution of $V_{00}$ becomes effectively a
continuous distribution, as seen for $\nu=1$. Once $\nu$ becomes very
large the GP has effectively learned the target function with very
little remaining uncertainty, and so the distribution of posterior
variances $V_{00}$ becomes increasingly peaked near zero; this trend
can be seen for $\nu=10$.

\Fref{fig:pop} (c) shows the equivalent plot for Erd\H{o}s-R\'enyi random graphs (see \sref{sec:results} for a description of this type of graph) with average degree 3.
Since graph structure is now no longer uniform, one does not see a
pattern of delta peaks for small $\nu$ as was the case for regular
graphs. There are however visible peaks, around $V_{00}=0$,
$V_{00}=1.15$ and $V_{00}=1.4$, superimposed on a continuous
distribution. This is from vertices with degree 1, 3
and 2 respectively. To 
see why, note that the cavity covariance matrices $\bm{V}$ can be
interpreted as messages from a vertex that are sent to a neighbour,
after incorporating the incoming messages from all other neighbours. Since
vertices with degree 1 have no 
incoming messages, all messages sent from such vertices will be
identical, resulting in a large peak in the distribution of $V_{00}$.
The same reason also gives peaks from vertices with higher degree,
as long as $\nu$ is small so that the majority of vertices has not yet seen
an example (corresponding to $\gamma=0$ in \eref{eqn:globalvarianceupdate}).
For example, the second peak in the figure is from vertices with
degree 3 that receive two incoming messages from neighbours with
degree 1. As these incoming messages are deterministic, so is the
outgoing message if no local example has been seen. The third peak
arises similarly from degree 2 vertices with an incoming message from
a vertex with degree 1.

\subsection{Local normalisation}\label{sec:local}

Using the insight gained from the simpler globally normalised kernel, we
now derive a prediction for the learning curve for the more
complicated case of a locally normalised kernel. Here $\kappa_i$ is equal to
the prior variance at vertex $i$, as calculated from the unnormalised
kernel. Although this may seem like a trivial 
change to the normalisation, it requires that properties of the
quenched disorder are effectively defined via a second thermal
average, making the calculation rather more complicated. The change
from global to local normalisation can also be shown to cause
significant changes to the GP prior over functions, and in turn to the
learning curves~\cite{UrryTBA}.

We begin our calculation by including the vertex-dependent normalisation
constants $\kappa_i$ in \eref{eqn:singlesite} via a
delta enforcement term 
\begin{equation}\label{eqn:localZdelta}
  \eqalign{
  \fl
  Z = \int
  \rmd\hpvecvec\rmd\bkappa\prod_{i\in\mathcal{V}}\exp\left(-\mathcal{H}(\hpvec_i,d_i,\kappa_i,\ni[i])\right)\prod_{(i,j)\in\mathcal{E}}\exp\left(-\mathcal{J}(\hpvec_i,\hpvec_j)\right)\\
  \times\prod_{i\in\mathcal{V}}\delta\left(\kappa_i - \frac{1}{Z\aux}\int \rmd\f\aux
  (f\aux[i])^2\exp\left(-\frac{1}{2}\f\aux\T\C_0^{-1}\f\aux\right)\right).}
\end{equation}
Here $Z\aux$ is defined to be the normaliser for an auxiliary GP
$\f\aux$, and $\C_0=\left(\bm{I} - a\bm{L}\right)^{p}$ is the
unnormalised kernel from \eref{eqn:rw}.

Because the definitions of the $\kappa_i$ depend on the entire graph in a non-local way, the partition function
\eref{eqn:localZdelta} no longer has the structure of a graphical model. Before we
can apply the replica method, we need to bring it back into this
form. This will require the introduction of an additional set of
replicas for the auxiliary GP.

Focussing on the $\kappa_i$-enforcement terms, we rewrite the delta
functions in terms of their Fourier transform by introducing a conjugate
variable $\hkappa_i$, and replace the integral over $\f\aux$ in the
exponent with an empirical average over $L\to\infty$ replicas of
$\f\aux$. With an element of foresight we also introduce $\lambda\aux\to0$
so that later equations are well behaved for $\hkappa_i=0$. The delta
enforcement terms from \eref{eqn:localZdelta} are then given by 
\begin{equation}
  \eqalign{
  \fl
  \lim_{L\to\infty\atop\lambda\aux\to0}\int \prod_{i\in\mathcal{V}} \frac{\rmd \hkappa_i}{2\pi}\prod_{l=1}^{L}\left(\frac{\rmd\f^l\aux}{Z\aux}\right)\exp\left(\sum_{i\in\mathcal{V}}\left[\rmi \hkappa_i\kappa_i
-\frac{1}{2}\left(\frac{2\rmi\hkappa_i}{L}+\lambda\aux\right)\sum_{l=1}^{L}(f\aux[i]^{l})^2\right]\right)
  \\\times
\exp\left(-\frac{1}{2}\sum_{l=1}^{L}(\f^l\aux)\T\C_0^{-1}\f^l\aux\right).}
\end{equation}
The non-local coupling via $\C_0^{-1}$ can be reduced to nearest
neighbour interactions by the same method we deployed for the original GP variables $\f$. 
In order to get \eref{eqn:localZdelta} into the form of a graphical
model it then remains to deal with the graph dependent $Z\aux$ term. We
bring $Z\aux$ into the numerator by introducing $m-1$
replicas of $Z\aux$ and taking $m\to0$ so that \mbox{$Z\aux^{-1} =
  \lim_{m\to0}Z\aux^{m-1}$}. As is conventional in replica
calculations, we then exchange the limits $m\to 0$ and $L\to\infty$,
taking the latter first. The $L$-replica
average over $(f\aux[i]^{l})^2 \equiv (f\aux[i]^{l,1})^{2}$ can then be
symmetrised to an $Lm$-replica average over the $(f^{l,b}\aux[i])^{2}$
with $b=1,\ldots,m$. This is because we have an infinite number $mL$
of replicas at any fixed $m>0$ so that both averages become
non-fluctuating. At this point the only effect of $m$ is via the total
number of replicas $mL$, and to simplify the notation we can fix
$m=1$. At the end of the calculation of $\langle \log 
Z\rangle$ we then in principle have to replace $L$ by $mL$ and take
$m\to 0$. However, since for the generalisation error we only need the
$\lambda$-derivative, which does not directly depend on the auxiliary degrees
of freedom, this will not be necessary.

With these simplifications, and after also rescaling $\hkappa_i$ by a
factor of $L$ for later convenience, the delta function constraints
for the $\kappa_i$ take the form 
\begin{equation}\label{eqn:mLreplicaZ}
\eqalign{
  \fl
  \lim_{\lambda\aux\to0 \atop L\to\infty}\int \prod_{i\in\mathcal{V}} \frac{\rmd\hkappa_iL}{2\pi}\prod_{l=1}^{L}\rmd\f^{l}\aux\exp\left(\sum_{i\in\mathcal{V}}\left[\rmi L\hkappa_i\kappa_i-\frac{1}{2}\left(2\rmi\hkappa_i+\lambda\aux\right)\sum_{l=1}^{L}(f\aux[i]^{l})^2\right]\right)\\
 \times\exp\left( -\frac{1}{2}\sum_{l=1}^{L}(\f^{l}\aux)\T\C_0^{-1}\f^{l}\aux\right).}
\end{equation}

Finally rewriting \eref{eqn:mLreplicaZ} using the same techniques as
we did for $Z$ in \sref{sec:partfunction}, by Fourier transforming the
covariance, integrating out the remaining terms, introducing
$2p$ additional variables at each site, rescaling according to $(h^{*})^q\aux[i]
= h^q\aux[i]d_{i}^{-1/2}$ and $(\hh^*)^q\aux[i] =
\hh^q\aux[i]d_{i}^{-1/2}$ and dropping the asterisks again, we arrive
at the desired graphical model form of the partition function: 
\begin{equation}
  \eqalign{
  \fl
  Z =\lim_{\lambda\aux\to0\atop L\to\infty}\int
  \rmd\hpvecvec\,\frac{\rmd\bkappa\rmd\bhkappa L^V}{(2\pi
  )^{V}}\prod_{l=1}^{L}\rmd\hpvecvec\aux^{l}\prod_{(i,j)\in\mathcal{E}}\exp\left(-\mathcal{J}(\hpvec_i,\hpvec_j) -
  \sum_{l=1}^{L}\mathcal{J}( \hpvec\aux[i]^{l},\hpvec\aux[j]^{l})\right)\\
  \times\prod_{i\in\mathcal{V}}\exp\left(-\mathcal{H}(\hpvec_i,d_i,\kappa_i,\ni[i])
  +\rmi L\kappa_i\hkappa_i-\sum_{l=1}^{L}\mathcal{H}\aux( \hpvec\aux[i]^{l},d_i,\hkappa_i)\right).}
  \end{equation}
Here we have defined
\begin{equation}
  \eqalign{
  \fl
  \mathcal{H}\aux( \hpvec\aux,d,\hkappa) = \frac{d}{2}\sum_{q=0}^{p}c_qh\aux^{0}h\aux^{q}
  +\frac{d}{2}\frac{( h\aux^{0})^2}{\lambda\aux +
  2\rmi\hkappa}-\rmi
d\sum_{q=1}^{p}\hh\aux^qh\aux^q\\-\frac{1}{2}\log\left(\frac{1}{\lambda\aux+2\rmi\hkappa}\right)\
.}
\end{equation}

With \eref{eqn:localZdelta} now in a form suitable for application of the
replica method we proceed initially as we did in the case of global
normalisation in \sref{sec:global}. We bring the graph and input
average inside the logarithm using the replica trick at the cost of
$n$ additional replicas. The resulting average of $Z^n$ reads as, 
\begin{equation}\label{eqn:Zlocalreplicatedsites}
  \eqalign{
  \fl
 \langle Z^{n}\rangle_{\x,\mathcal{G}} = \lim_{L\to\infty}\left\langle\int
  \prod_{a=1}^{n}\left(\rmd\hpvecvec^{a}\,\frac{\rmd\bkappa^a\rmd\bhkappa^a L^V}{(2\pi )^{V}}\prod_{l= 1}^{L}\rmd\hpvecvec\aux^{l,a}\right)\prod_{i\in\mathcal{V}}\exp\left(-\sum_{a=1}^{n}\mathcal{H}(\hpvec^a_i,d_i,\kappa_i,\ni[i])\right.\right.\\
  \left.+\rmi L\sum_{a=1}^{n}\kappa^{a}_i\hkappa^{a}_i-\sum_{a=1}^{n}\sum_{l=1}^{L}
  \mathcal{H}\aux(\hpvec\aux[i]^{l,a},d_i,\hkappa^a_i)\right)\\
  \left.\times\prod_{(i,j)\in\mathcal{E}}\exp\left(-\sum_{a=1}^{n}\mathcal{J}(\hpvec^a_i,\hpvec^a_j) -
  \sum_{a=1}^{n}\sum_{l=1}^{L}\mathcal{J}(\hpvec\aux[i]^{l,a},\hpvec\aux[j]^{l,a})\right)\right\rangle_{\ni,\mathcal{G}}.}
\end{equation}

Similarly to the global case we introduce replica densities
\begin{equation}
\eqalign{
  \fl
\rholocfunc\\
=\frac{1}{V}\sum_i\rme^{\rmi\hd_i}\prod_{a=1}^{n}\delta(\hpvec^a-\hpvec^a_i)\delta(\kappa^a-\kappa_i^a)\delta(\hkappa^a-\hkappa_i^a)\prod_{l=1}^{L}\delta(\hpvec\aux^{l,a}-\hpvec\aux[i]^{l,a})
}
\end{equation}
with enforcement via conjugate densities $\hrho$. We substitute the
densities into \eref{eqn:Zlocalreplicatedsites} and perform the
average over the graph ensemble to derive, using techniques similar to
those for the global case (see \ref{app:globalpresaddle}), 
\begin{equation}\label{eqn:localZpresaddle}
  \langle Z^{n}\rangle_{\x,\mathcal{G}} =\frac{1}{\mathcal{N}} \int \mathcal{D}\rho \mathcal{D}\hrho 
  \exp\left(V\left[\frac{\bar{d}}{2}(S_{1}[\rho]-1) - \rmi S_{2}[\rho,\hrho] +
  S_{3}[\hrho]\right]\right),
\end{equation}
where 
\numparts
\begin{eqnarray}
  \eqalign{
  \fl
  S_1[\rho] = \int \prod_{a=1}^{n}\left(\rmd\hpvec^a
  \rmd(\hpvec^a)'\frac{\rmd\kappa^a\rmd\hkappa^aL}{2\pi }\frac{\rmd(\kappa^a)'\rmd(\hkappa^a)'L}{2\pi }\prod_{l=1}^{L}\rmd\hpvec\aux^{l,a}\rmd(\hpvec\aux^{l,a})'\right)\rhofuncshort\rhofuncshortdash\\
  \times\exp\left(-\sum_{a=1}^{n}\mathcal{J}(\hpvec^a,(\hpvec^a)')-
  \sum_{a=1}^{n}\sum_{l=1}^{L}\mathcal{J}(\hpvec\aux^{l,a},(\hpvec\aux^{l,a})')\right)}\label{eqn:localS1}\\
  \fl S_{2}[\rho,\hrho] = \int\prod_{a=1}^{n}\left(\rmd\hpvec^a \frac{\rmd\kappa^a\rmd\hkappa^aL}{2\pi }\prod_{l=1}^{L}\rmd\hpvec\aux^{l,a}\right)\rhofuncshort\hrhofuncshort\label{eqn:localS2}\\
  \eqalign{
  \fl S_{3}[\hrho] =
  \sum_{d}p(d)\log\left\langle\int\prod_{a=1}^{n}\left(\rmd\hpvec^a\frac{\rmd\kappa^a\rmd\hkappa^aL}{2\pi }\prod_{l=1}^{L}\rmd\hpvec\aux^{l,a}\right)\exp\left(-\sum_{a=1}^{n}\mathcal{H}(\hpvec^a,d,\kappa^a,\ni)\right.\right.\\
  \left.\left. +\rmi L\sum_{a=1}^{n}\kappa^{a}\hkappa^a - \sum_{a=1}^{n}\sum_{l=1}^{L}
  \mathcal{H}\aux(\hpvec\aux^{l,a},d,\hkappa^a)\right)
  \frac{\left(\rmi\hrhofuncshort\right)^d}{d!}\right\rangle_{\gamma}.}\label{eqn:localS3}
\end{eqnarray}
\endnumparts
We have not written the limits $\lambda\aux\to 0$ and $L\to\infty$
explicitly here.

As before, the expression \eref{eqn:localZpresaddle} for the average
of the replicated partition function will be dominated by its saddle
point for large $V$. We assume that this saddle point has a replica
symmetric form and set
\begin{eqnarray}
\rhofuncshort = \int
\mathcal{D}\psi\,\pi[\psi]\prod_{a=1}^{n}\frac{\exp\left(-\psirenormfunc[a]\right)}{Z[\psi]}\label{eqn:localrho}\\
\hrhofuncshort = -\rmi\bar{d}\int
\mathcal{D}\hpsi\,\hpi[\hpsi]\prod_{a=1}^{n}\frac{\exp\left(-\hpsirenormfunc[a]\right)}{Z[\hpsi]}.\label{eqn:localhrho}
\end{eqnarray}

Substituting \eref{eqn:localrho} and \eref{eqn:localhrho} into
\eref{eqn:localS1} to \eref{eqn:localS3},
we can write each of the exponent functions as expansions in $n$, 
setting $S_{i} = S_{i}^{O(1)} + nS_{i}^{O(n)}$ up to linear order in $n$. 
As before, and as is standard in replica calculations, the subleading $O(n)$ terms are
required in order to calculate the generalisation error. The novel feature of this calculation
is the dependence on $L$ replicas of the auxiliary variables. As we will show, this leads to equations 
that couple the
$\hpvec$ to the $\{\hpvec\aux^{l}\}$ via the covariance matrix of the latter.

One finds that the expansions for $S_{1}$, $S_{2}$ and $S_{3}$ can be calculated in a similar manner to the global case (see \ref{app:globalupdate}). Leading $O(n^0)$ terms once more cancel the normaliser $\mathcal{N}$ at the saddle point, and we are left with optimising $O(n)$ terms given by
\begin{eqnarray}
\eqalign{
  \fl S_{1}^{O(n)}[\rho] = \int
  \mathcal{D}\psi\,\pi[\psi]\int\mathcal{D}\psi'\,\pi[\psi']\log\left[\int\rmd\hpvec\frac{\rmd\kappa\rmd\hkappa L}{2\pi}\rmd\hpvec'\frac{\rmd\kappa'\rmd\hkappa' L}{2\pi}\prod_{l}\rmd\hpvec\aux^l\rmd(\hpvec\aux^l)'\right.\\\exp\left(-\psirenormfunc-\psirenormfuncdash\right)
  \\
  \times\left.\frac{\exp\left(-\mathcal{J}(\hpvec,\hpvec')-\sum_{l}\mathcal{J}(\hpvec\aux^l,(\hpvec\aux^l)')\right)}{Z[\psi]Z[\psi']}\right]}\label{eqn:localS1On}\\
  \eqalign{
  \fl S_{2}^{O(n)}[\rho,\hrho] =
  -\rmi\bar{d}\int\mathcal{D}\psi\,\pi[\psi]\int\mathcal{D}\hpsi\,\hpi[\hpsi]\log\left[\int\rmd\hpvec\frac{\rmd\kappa\rmd\hkappa L}{2\pi}\prod_l\rmd\hpvec\aux^l\right.\\
\left.  \frac{\exp\left(-\psirenormfunc-\hpsirenormfunc[][]\right)}{Z[\psi]Z[\hpsi]}\right]}\label{eqn:localS2On}\\
\eqalign{
  \fl S_{3}^{O(n)}[\hrho] = \sum_{d}p(d)\left\langle\int
  \prod_{i=1}^{d}\mathcal{D}\hpsi^{i}\,\hpi[\hpsi^{i}]\log\left[\int\rmd\hpvec\frac{\rmd\kappa\rmd\hkappa L}{2\pi}\prod_l\rmd\hpvec\aux^l  \right.\right.\\
\exp\left(-\mathcal{H}(\hpvec,d,\kappa,\ni)+iL\kappa\hkappa - \sum_{l}\mathcal{H}\aux(\hpvec\aux^{l},d,\hkappa)\right)
\\
  \left.\left.
  \times  \frac{\exp\left(-\sum_{i=1}^{d}\hpsirenormfunc[][i]\right)}{\prod_{i=1}^{d}Z[\hpsi^{i}]}\right]\right\rangle_{\ni},\label{eqn:localS3Onpresaddle}}
\end{eqnarray}
subject to $\pi$ and $\hpi$ being normalised density functions.

Proceeding as we did in the global case (see \ref{app:globalupdate}) we see that the saddle point conditions with respect to $\pi$ and $\hpi$ can be calculated as
\begin{equation}\label{eqn:localupdatepsi}
\pi[\psi] = \sum_{d}\frac{p(d)d}{\bar{d}}\int \prod_{i=1}^{d-1}\mathcal{D}\psi^i\,\pi[\psi^i]\Bigg\langle\delta(\psi - \Psi[\hPsi[\psi^{1}],\ldots,\hPsi[\psi^{d-1}]])\Bigg\rangle_{\ni},
\end{equation}
with
\begin{eqnarray}
\eqalign
{
\fl \Psi[\hpsi^{1},\ldots,\hpsi^{d-1}] = \sum_{i=1}^{d-1}\hpsirenormfunc[][i] - \rmi L \kappa\hkappa+ \mathcal{H}(\hpvec,d,\kappa,\ni)\\
+\sum_l\mathcal{H}\aux(\hpvec\aux^l,d,\hkappa),\label{eqn:localPsi}}\\
\eqalign{
\fl
\hPsi[\psi'] = -\log\int\rmd \hpvec'\frac{\rmd\kappa'\rmd\hkappa' L}{2\pi} \prod_l \rmd(\hpvec\aux^l)' \exp\Bigg(-\psirenormfuncdash \Bigg)\\
\times\exp\Bigg(- \mathcal{J}(\hpvec,\hpvec')-\sum_l\mathcal{J}(\hpvec\aux^l,(\hpvec\aux^l)')\Bigg).
\label{eqn:localhPsi}}
\end{eqnarray}

Equations \eref{eqn:localupdatepsi}, \eref{eqn:localPsi} and \eref{eqn:localhPsi} are just the analogues of the global saddle point conditions \eref{eqn:globalupdate}, \eref{eqn:globalPsi} and \eref{eqn:globalhPsi} respectively, for a larger set of variables. Similarly an analogous expression for the generalisation error can be calculated as (see \ref{app:globalwoodburyderivations} for the global equivalent),
\begin{equation}\label{eqn:locallcgenL}
\eqalign{
\fl\epsilon_g = \lim_{\lambda\to 0}\sum_{d}p(d)\left\langle \int \prod_{i=1}^{d}\mathcal{D}\hpsi^{i}\hpi[\hpsi^{i}]\int \rmd \hpvec \frac{\rmd\kappa\rmd\hkappa L}{2\pi}\prod_{l=1}^{L}\rmd\hpvec\aux^l\right.\\
\left.\left(\frac{1}{\gamma/\sigma^2 + \lambda} - \frac{\kappa d\e0\T\hpvec\hpvec\T\e0}{(\gamma/\sigma^{2}+\lambda)^{2}}\right)q(\hpvec,\kappa,\hkappa,\{\hpvec\aux^{l}\})\right\rangle_{\ni},
}
\end{equation}
with
\begin{equation}\label{eqn:locallcpsi}
q(\hpvec,\kappa,\hkappa,\{\hpvec\aux^{l}\}) \propto \exp\left(-\Psi[\hPsi[\psi^{1}],\ldots,\hPsi[\psi^{d}]](\hpvec,\kappa,\hkappa,\{\hpvec\aux^{l}\})\right),
\end{equation}
and $\e0$ defined as before as $\e0=(1,0,\ldots,0)\T$. The proportionality factor in \eref{eqn:locallcpsi} is determined so that $q$ is normalised with respect to integration over
$\hpvec$, $\hpvec\aux$, $\kappa$ and $\hkappa L/(2\pi)$.

While the replica calculation for local normalisation has so far broadly followed that of the global case in \sref{sec:global}, we now need to take some additional steps to deal with the fact that the number of auxiliary variables diverges in the desired limit $L\to\infty$. Fortunately this limit allows us to make simplifications in both the saddle point condition \eref{eqn:localupdatepsi} and the expression \eref{eqn:locallcgenL} for the generalisation error. By using large $L$-saddle point methods we will obtain a numerically tractable set of equations from which we will be able to predict the generalisation error, \eref{eqn:epgdef}, for any number of training examples $N$.

Closer inspection of \eref{eqn:localhPsi} shows that $\hPsi[\psi]$ 
only depends on the marginal $-\log \int \rmd\kappa\rmd\hat\kappa L/(2\pi) \exp(-\psi(\hpvec,\kappa,\hkappa,\{\hpvec\aux^{l}\}))$; let us denote this by $\phi(\hpvec,\{\hpvec\aux^{l}\})$. $\hPsi[\psi]$ is also itself independent of $\kappa$ and $\hkappa$, and to emphasize this we will write the function it produces as $\hat\phi$ and the functional itself as $\hPhi[\phi]$.
%
%
In the limit of large $L$, the relevant marginalised (over $\kappa$ and $\hkappa$) versions of \eref{eqn:localPsi} and \eref{eqn:localhPsi} can now be calculated by a saddle point evaluation (see \ref{app:localO(n)derive}) to give
\begin{eqnarray}
\eqalign{
\fl\Phi[\hphi^{1},\ldots,\hphi^{d-1}](\hpvec,\{\hpvec\aux^{l}\}) = \sum_{i=1}^{d-1}\hphi^i(\hpvec,\{\hpvec\aux^{l}\}) + \mathcal{H}\Big(\hpvec,d,
v\Big(\frac{1}{L}\sum_{l}\hpvec\aux^{l}(\hpvec\aux^{l})\T\Big),\ni\Big)\\
+\sum_l\mathcal{H}\aux(\hpvec\aux^l,d,0),
\label{eqn:locallargeLPhi}}\\
\eqalign{
\fl \hPhi[\phi'] = -\log\int\rmd \hpvec' \prod_l \rmd(\hpvec\aux^l)' \exp\Bigg(-\phifuncdash - \mathcal{J}(\hpvec,\hpvec')\Bigg)\\
\times\exp\Bigg(- \sum_l\mathcal{J}(\hpvec\aux^l,(\hpvec\aux^l)')\Bigg).
\label{eqn:locallargeLhPhi}}
\end{eqnarray}
Here we have defined
\begin{equation}\label{eqn:localauxvar}
 v(\Hmat)= 
\frac{\lambda\aux-d\e0\T\Hmat\e0}{\lambda\aux^2}.
\end{equation}
as a function of the empirical auxiliary covariance $\Hmat=\frac{1}{L}\sum_{l}\hpvec\aux^l(\hpvec\aux^l)\T$.
Combining \eref{eqn:locallargeLPhi} and \eref{eqn:locallargeLhPhi} the $\kappa$ and $\hkappa$-independent version of \eref{eqn:localupdatepsi} is then the self-consistency condition
\begin{equation}\label{eqn:localupdatephi}
\pi_{\phi}[\phi] = \sum_{d}\frac{p(d)d}{\bar{d}}\int \prod_{i=1}^{d-1}\mathcal{D}\phi^i\,\pi_{\phi}[\phi^i]\Bigg\langle\delta\bigg(\phi - \Phi[\hPhi[\phi^{1}],\ldots,\hPhi[\phi^{d-1}]]\bigg)\Bigg\rangle_{\ni},
\end{equation}
for the distribution $\pi_\phi$ of the functions $\phi$. 
Similar large-$L$ saddle point arguments yield for the generalisation error \eref{eqn:locallcpsi}
\begin{equation}\label{eqn:locallclargeLpsi}
\eqalign{
\fl\epsilon_g = \lim_{\lambda\to 0}\sum_{d}p(d)\left\langle \int \prod_{i=1}^{d}\mathcal{D}\hphi^{i}\hpi_{\phi}[\hphi^{i}]\int \rmd \hpvec \prod_{l=1}^{L}\rmd\hpvec\aux^l\right.\\
\left.\left(\frac{1}{\gamma/\sigma^2 + \lambda} - \frac{v\Big(\frac{1}{L}\sum_{l}\hpvec\aux^{l}(\hpvec\aux^{l})\T\Big)d\e0\T\hpvec\hpvec\T\e0}{(\gamma/\sigma^{2}+\lambda)^{2}}\right)q_{\phi}(\hpvec,\{\hpvec\aux^{l}\})\right\rangle_{\ni},
}
\end{equation}
with
\begin{equation}\label{eqn:qphi}
q_{\phi}(\hpvec,\{\hpvec\aux^{l}\}) \propto \exp\left(-\Phi[\hPhi[\phi^{1}],\ldots,\hPhi[\phi^{d}]](\hpvec,\{\hpvec\aux^{l}\})\right).
\end{equation}

With the large-$L$ limit taken, an ansatz for solving the saddle point conditions \eref{eqn:localupdatephi} can now be proposed. We take the `energy functions' $\phiindeprenormfunc$ to be of quadratic form again. In contrast to the case of global kernel normalisation, however, the covariances of the $\hpvec$ variables are chosen to depend on the local $\hpvec\aux$ variables through the empirical auxiliary covariance $\Hmat=\frac{1}{L}\sum_{l}\hpvec\aux^l(\hpvec\aux^l)\T$. This dependence is motivated by the definitions \eref{eqn:locallargeLPhi}, \eref{eqn:locallargeLhPhi}, \eref{eqn:localauxvar} and \eref{eqn:qphi} in which interactions between $\hpvec$ and the auxiliary variables only appear via $\Hmat$. Specifically, we write
\begin{equation}\label{eqn:localpsiansatz}
\phiindeprenormfunc = \frac{1}{2}\hpvec\T [\localfunc(\Hmat)]^{-1}\hpvec +\frac{1}{2}\sum_{l}(\hpvec^l\aux)\T\V\aux^{-1}\hpvec^l\aux,
\end{equation}
for $\localfunc(\Hmat)$ a matrix function of the empirical covariance $\Hmat$ of the $\hpvec\aux$ variables. 

Substituting \eref{eqn:localpsiansatz} into \eref{eqn:locallargeLhPhi} we have
\begin{equation}\label{eqn:hPhi}
\eqalign{
\fl\hPhi[\phi'] = -\log \int \rmd \hpvec' \prod_{l=1}^{L}\rmd (\hpvec\aux^l)' \exp\Big(- \frac{1}{2}(\hpvec')\T [\localfunc'\left(\Hmat'\right)]^{-1}\hpvec' - \hpvec\T\X\hpvec'\\ -\frac{1}{2}\sum_{l}(\hpvec\aux^{l})'{}\T(\V'\aux)^{-1}(\hpvec\aux^{l})'-\sum_{l}(\hpvec^{l}\aux)\T\X(\hpvec^{l}\aux)'\Big).
}
\end{equation}
We wish to integrate \eref{eqn:hPhi} but a direct attack is complicated because of the $\hpvec\aux'$-dependence in $\localfunc'$. However, we can think of the factors in the second line of \eref{eqn:hPhi} as defining a Gaussian weight on $\Delta^l \equiv (\hpvec^{l}\aux)'$, so that the $\Delta^l$ are $L$ independent and identically distributed Gaussian random variables with distribution $\Delta^l\sim\mathcal{N}(-\V\aux'\X\hpvec^l\aux,\V\aux')$. Discarding irrelevant additive constants yields then
\begin{equation}\label{eqn:hPhiDelta}
\eqalign{
\fl\hPhi[\phi'] = -\log \Biggl[
\exp\Big(\frac{1}{2}\sum_{l}(\hpvec^l\aux)\T\X\V'\aux\X\hpvec^l\aux\Big)
\int \rmd \hpvec'\exp(- \hpvec\T\X\hpvec')\\\left\langle  \exp\Bigg(- \frac{1}{2}(\hpvec')\T \Big[\localfunc'\Big(\frac{1}{L}\sum_{l}\Delta^l(\Delta^l)\T\Big)\Big]^{-1}\hpvec'\Bigg)\right\rangle_{\Delta^l}\Biggr].
}
\end{equation}
The key step is now that, in the limit of large $L$, the argument of $\localfunc'$ becomes self averaging, so that we may rewrite \eref{eqn:hPhiDelta} as
\begin{equation}\label{eqn:hPhifinal}
\fl\hPhi[\phi'] =-\frac{1}{2}\sum_{l}(\hpvec^l\aux)\T\X\V'\aux\X\hpvec^l\aux
- \frac{1}{2}\hpvec\T\X\localfunc'(\V\aux' + \V\aux'\X\Hmat\X\V\aux')\X\hpvec.
\end{equation}

With the integration in $\hPhi$ performed we may substitute \eref{eqn:localpsiansatz} and \eref{eqn:hPhifinal} into \eref{eqn:localupdatephi} to get the saddle point condition
\begin{equation}\label{eqn:localnosolveupdate}
\eqalign{
\fl
\pi[\V\aux,\localfunc(\,\cdot\,)] = \sum_{d}\frac{p(d)d}{\bar{d}}\int \prod_{i=1}^{d-1}\rmd\V\aux^i\mathcal{D}\localfunc^i\pi[\V\aux^i,\localfunc^i(\,\cdot\,)]\\
\left\langle\prod_{\Hmat}\delta\Big(\localfunc(\Hmat)-
[\tilde{\O}(\Hmat)-\sum_{i=1}^{d-1}\X\localfunc^{i}(\V\aux^i + \V\aux^i\X\Hmat\X\V\aux^i)\X]^{-1}\Big)\right.\\
\left.\times\delta\Big(\V\aux-[\O\aux-\sum_{i=1}^{d-1}\X\V^i\aux\X]^{-1}\Big)\right\rangle_{\ni},
}
\end{equation}
where we have defined $\O\aux$ in a similar manner to $\O$ (see \eref{eqn:globalOX}) but with $\kappa=1$ and $\gamma=0$, and
\begin{equation}
 \fl \tilde{\O}(\Hmat) = d\left(\begin{array}{cccc|ccc}
c_0 \!+\!\frac{1}{dv(\Hmat)(\gamma/\sigma^{2} +\lambda)} &
\frac{1}{2}c_1 & \dots & \frac{1}{2}c_{p} &
0 & \dots & 0 \\
\frac{1}{2}c_{1}& & & &
-\rmi & & \\
\vdots & & & &
 & \ddots & \\
\frac{1}{2}c_{p}& & & &
 & & -\rmi\\[0.5mm]
\hline
0 & -\rmi & & &
 & & \\
\vdots & & \ddots & &
 & \bm{0}_{p,p} & \\
0 & & & -\rmi & 
 & &
\end{array}\right).
\end{equation}
The $\prod_{\Hmat}$ in \eref{eqn:localnosolveupdate} represents the definition of the matrix function $\localfunc(\Hmat)$ for all possible arguments $\Hmat$.

A similar large-$L$ self-averaging argument can be made for the generalisation error, \eref{eqn:locallclargeLpsi}. Since each $\hpvec\aux^l$ belongs to a Gaussian distribution with common covariance given by $q_\phi$ we may replace the empirical covariances $\Hmat=\frac{1}{L}\sum_{l}\hpvec\aux^{l}(\hpvec\aux^{l})\T$ in \eref{eqn:locallclargeLpsi} with $\Hmat_{m}=(\O\aux - \sum_{i=1}^{d}\X\V^{i}\aux\X)^{-1}$, the local auxiliary marginal. This simplifies \eref{eqn:locallclargeLpsi} to give
\begin{equation}\label{eqn:locallearningcurveempirical}
\eqalign{
\fl\epsilon_g = \lim_{\lambda\to 0}\sum_{d}p(d)\left\langle \int \prod_{i=1}^{d}\rmd\V\aux^{i}\rmd\localfunc^{i}(\,\cdot\,)\pi[\V\aux^{i},\localfunc^{i}(\,\cdot\,)]\right.\\
\left.\left(\frac{1}{\gamma/\sigma^2 + \lambda} - \frac{v(\Hmat_{m})d\e0\T\V_m\e0}{(\gamma/\sigma^{2}+\lambda)^{2}}\right)\right\rangle_{\ni},
}
\end{equation}
with
\begin{equation}
\V_m = \Big[\tilde{\O}(\Hmat_{m}) - \sum_{i=1}^{d}\X\localfunc^{i}(\V\aux^{i}+\V\aux^{i}\X\Hmat_{m}\X\V\aux^{i})\X\Big]^{-1}.
\end{equation}
It is worth comparing \eref{eqn:locallearningcurveempirical} to the corresponding expression \eref{eqn:globalsingularlearningcurve} for the case of a globally normalised kernel. The normalisation factor $\kappa$ there has been replaced by
\begin{equation}\label{eqn:vHmat}
v(\Hmat_{m}) = \frac{\lambda\aux-d\e0\T\Hmat_{m}\e0}{\lambda\aux^2}.
\end{equation}
This itself again looks like a contribution to a generalisation error, but with $\gamma=0$ and $\kappa=1$, which is reassuring: for $\gamma=0$ (no training examples), the generalisation error is just the prior variance. So $v(\Hmat_{m})$ is the local prior variance of the auxiliary variables, and it is this quantity that should provide the normalisation factor for the original variables.

The inverse matrix in \eref{eqn:globalsingularlearningcurve} corresponds to $\V_m$ in \eref{eqn:locallearningcurveempirical} and in both cases is the covariance matrix of $\hpvec$. In the locally normalised scenario, the large $L$-limit tells us that the matrix function $\localfunc(\,\cdot\,)$ is here evaluated at a definite point, namely the local auxiliary marginal, $\Hmat_m$.

Conceptually, we are now in principle done: we have taken $L\to\infty$ and have a self-consistency condition for the distribution of $\V\aux,\localfunc(\,\cdot\,)$. But we have paid for this by the fact that $\localfunc(\,\cdot\,)$ is an entire matrix function. This would be very difficult to parameterise for a numerical solution of \eref{eqn:localnosolveupdate} by population dynamics. Our final step in the analysis is therefore to reduce the description to one in terms of matrices rather than matrix functions.

To motivate this, we recall that
since \eref{eqn:localpsiansatz} amounts to assuming replica symmetry, \eref{eqn:localnosolveupdate} has a cavity interpretation~\cite{Mezard1987}. For a vertex with degree $d$ we sample $d-1$ neighbours independently and combine their cavity distributions with information about the current vertex to create a cavity distribution or `message' to send to the $d$-th neighbour. In the case of local normalisation, the $d-1$ neighbours each pass an auxiliary cavity covariance $\V\aux$ and a cavity covariance function $\localfunc(\Hmat)$. Auxiliary covariances are combined with local information to create a new auxiliary covariance according to $\V\aux=(\O\aux-\sum_{i=1}^{d-1}\X\V^i\aux\X)^{-1}$, in direct analogy to \eref{eqn:globalvarianceupdate} for globally normalised kernels. The covariance function $\localfunc(\Hmat)$ is calculated slightly differently, as specified by the functional delta function in the second line of \eref{eqn:localnosolveupdate}, due to its dependence on the auxiliary covariance through its argument. It combines incoming function covariances $\localfunc^{i}$ evaluated at $\V\aux^{i}+\V\aux^{i}\X\Hmat\X\V\aux^{i}$ with local information dependent on the empirical covariance $\Hmat$ to calculate the outgoing cavity covariance function $\localfunc(\Hmat)$.

The important point is now that when we come to calculate the generalisation error, we do not require the full functions $\tilde{\O}(\,\cdot\,)$ and $\localfunc^i(\,\cdot\,)$, but only their values at specific arguments. The marginal covariance in $\tilde{\O}(\Hmat_m)$ can be written as
\begin{equation}
\label{eq:Hmat}
\Hmat_m = (\O\aux - \sum_{i=1}^{d}\X\V^{i}\aux\X)^{-1} = (\V\aux^{-1} - 
\X\rev{\V}[\auxrm][]\X)^{-1},
\end{equation}
where $\V\aux=(\O\aux-\sum_{i=1}^{d-1}\X\V^i\aux\X)^{-1}$ is the auxiliary covariance message that would be sent to the $d$-th neighbour, and $\rev{\V}[\auxrm]\equiv \V^d\aux$ is the {\em reverse message} that this vertex sends. This form of argument is maintained consistently at neighbouring vertices in the self-consistency condition \eref{eqn:localnosolveupdate}. Indeed, substituting $\Hmat_{m}$ into the argument of $\localfunc^i$ on the right hand side of \eref{eqn:localnosolveupdate} we see that, by application of Woodbury's identity, this may be rewritten as
\begin{equation}\label{eqn:localfuncarg}
\eqalign{
\fl
\V\aux^{i} + \V\aux^{i}\X\Hmat_{m}\X\V\aux^{i} = \V\aux^{i} + \V\aux^{i}\X\Big(\O\aux - \sum_{j=1}^{d-1}\X\V\aux^{j}\X - \X\rev{\V}[\auxrm]\X\Big)^{-1}\X\V\aux^{i}\\
= \Big[(\V\aux^{i})^{-1} -\X\Big(\O\aux-\sum_{j\neq i}\X\V\aux^{j}\X - \X\rev{\V}[\auxrm]\X\Big)^{-1}\X\Big]^{-1}.
}
\end{equation}
This is of the same form as \eref{eq:Hmat}, and has the analogous interpretation of the marginal auxiliary covariance at vertex $i$, $\Hmat^i_m$. It is expressed again in terms of a message this vertex sends, $\V\aux^i$, and a reverse message sent to this vertex. The latter is $\rev{\V}[\auxrm][i] = (\O\aux-\sum_{j\neq i}\X\V\aux^{j}\X - \X\rev{\V}[\auxrm]\X)^{-1}$ and is constructed from information received from all other neighbours of the central vertex, including the $d$-th one.

The discussion so far suggests that we should consider a reduced but conditional distribution of messages where the covariance function $\localfunc(\,\cdot\,)$ is evaluated only at the local marginal auxiliary covariance:
\begin{equation}\label{eqn:pinofunc}
\fl\pi[\V\aux,\V|\rev{\V}[\auxrm]] = \int \mathcal{D}\localfunc \pi[\V\aux,\localfunc(\,\cdot\,)] \delta\bigg(\V-\localfunc([\V\aux^{-1}-\X\rev{\V}[\auxrm]\X]^{-1})\bigg).
\end{equation}
Substituting this into \eref{eqn:localnosolveupdate}, we can now solve the problem of passing the full functions $\localfunc$. After some algebra, including applying the trick \eref{eqn:localfuncarg}, we get the following self-consistency equation:
\begin{equation}\label{eqn:localupdatecomplicated}
\eqalign{
\fl
\pi[\V\aux,\V|\rev{\V}[\auxrm]] = \sum_{d}\frac{p(d)d}{\bar{d}}\int \prod_{i=1}^{d-1}\rmd\V\aux^i\rmd\V^i\rmd\rev{\V}[\auxrm][i]
\,\pi[\V\aux^i,\V^i|\rev{\V}[\auxrm][i]]\\
\left\langle\delta\Big(\V\aux-\Big[\O\aux-\sum_{i=1}^{d-1}\X\V^i\aux\X\Big]^{-1}\Big)\delta\Big(\V-\Big[\tilde{\O}-\sum_{i=1}^{d-1}\X\V^i\X\Big]^{-1}\Big)\right.\\
\left.\times\prod_{i=1}^{d-1}\delta\Big(\rev{\V}[\auxrm][i]-\Big[\O\aux-\sum_{j\neq i}\X\V\aux^{j}\X - \X\rev{\V}[\auxrm]\X\Big]^{-1}\Big)\right\rangle_{\ni}
.}
\end{equation}
where we have abbreviated $\tilde{\O} = \tilde{\O}([\V\aux^{-1}-\X\rev{\V}[\auxrm]\X]^{-1})$. The generalisation error \eref{eqn:locallearningcurveempirical} can be rewritten in the same fashion as
\begin{equation}\label{eqn:locallearningcurveempirical2}
\eqalign{
\fl\epsilon_g = \lim_{\lambda\to 0}\sum_{d}p(d)\left\langle \int \prod_{i=1}^{d}\rmd\V\aux^{i}
\rmd\V^i\rmd\rev{\V}[\auxrm][i]\,\pi[\V\aux^i,\V^i|\rev{\V}[\auxrm][i]]\right.\\
\left.\left(\frac{1}{\gamma/\sigma^2 + \lambda} - \frac{v(\Hmat_{m})d\e0\T\V_m\e0}{(\gamma/\sigma^{2}+\lambda)^{2}}\right)\right.
\\
\left.\times
\prod_{i=1}^{d}\delta\Big(\rev{\V}[\auxrm][i]-\Big[\O\aux-\sum_{j\neq i}\X\V\aux^{j}\X - \X\rev{\V}[\auxrm]\X\Big]^{-1}\Big)
\right\rangle_{\ni}.
}
\end{equation}
Both here and in the self-consistency equation \eref{eqn:localupdatecomplicated}, the delta functions for the reverse messages effectively ensure that these messages are consistent with the forward messages $\V\aux^i$ and with $\rev{\V}[\auxrm]$.

To find $\pi[\V\aux,\V|\rev{\V}[\auxrm]]$ numerically from \eref{eqn:localupdatecomplicated} would require conditional population dynamics (see \cite{FontClosTBA}), which is still rather challenging. We therefore now make an approximation. Since $\rev{\V}[\auxrm]$ is a message from the $d$-th neighbour like the other auxiliary covariances $\V\aux^i$ being sent from neighbours $i=1,\ldots,d-1$, it has probability weight $\pi[\rev{\V}[\auxrm]]$. Multiplying \eref{eqn:localupdatecomplicated} by this weight and integrating over $\rev{\V}[\auxrm]$ gives on the left hand side the unconditional distribution $\pi[\V\aux,\V]$. Our approximation consists of dropping at this stage the conditioning also on the right hand side, replacing $\pi[\V\aux^i,\V^i|\rev{\V}[\auxrm][i]]$ by $\pi[\V\aux^i,\V^i]$. The reverse messages can then be integrated out, leading to the approximate self-consistency equation
\begin{equation}\label{eqn:localupdatenofunc}
\eqalign{
\fl
\pi[\V\aux,\V] = \sum_{d}\frac{p(d)d}{\bar{d}}\int \prod_{i=1}^{d-1}\rmd\V\aux^i\rmd\V^i\pi[\V\aux^i,\V^i]\int \rmd\rev{\V}[\auxrm]\pi[\rev{\V}[\auxrm]]\\
\left\langle\delta\Big(\V\aux-\Big[\O\aux-\sum_{i=1}^{d-1}\X\V^i\aux\X\Big]^{-1}\Big)\delta\Big(\V-
\Big[\tilde{\O}-\sum_{i=1}^{d-1}\X\V^i\X\Big]^{-1}\Big)\right\rangle_{\ni}.
}
\end{equation}
\Eref{eqn:localupdatenofunc} has two apparent singularities, one in $\tilde{\O}$ as $\lambda\to 0$, if $\ni=0$, and another one in $v(\Hmat_m)$ -- which appears in the definition of $\tilde{\O}$ -- for $\lambda\aux\to 0$. Apparent divergences in $\tilde{\O}$ caused by $\lambda\to 0$ are resolved in a similar manner to those in \sref{sec:global}. Similar problems in $v(\Hmat_m)$ as $\lambda\aux\to 0$ can be avoided by introducing $\M\aux[d] = \O\aux - \sum_{i=1}^{d}\X\V\aux^{i}\X -
\e0\frac{d}{\lambda\aux}\e0\T$ and applying the Woodbury identity (see \ref{app:globalwoodburyderivations}) to give $v(\Hmat_m) = (d\e0\T\M\aux[d]^{-1}\e0)^{-1}$ in the limit $\lambda\aux\to 0$.

The approximation to the generalisation error \eref{eqn:locallearningcurveempirical2} that corresponds to the approximate self-consistency equation \eref{eqn:localupdatenofunc} reads, after again applying the Woodbury formula to deal with the two apparent divergences for $\lambda\to 0$ and $\lambda\aux\to 0$,
\begin{equation}\label{eqn:locallearningcurveempirical3}
\eqalign{
\fl\epsilon_g = \sum_{d}p(d)\left\langle \int \prod_{i=1}^{d}\rmd\V\aux^{i}
\rmd\V^i\,\pi[\V\aux^i,\V^i]
\frac{1}{\gamma/\sigma^2 + (\e0\T\M\aux[d]^{-1}\e0)^{-1}\e0\T\M_{d}^{-1}\e0}\right\rangle_{\ni}
}
\end{equation}
with $\M_{d}$ defined in a similar manner to \eref{eqn:woodbury}. \Eref{eqn:locallearningcurveempirical3} can be evaluated straightforwardly once a population estimate for $\pi[\V\aux,\V]$ has been found. The interpretation of \eref{eqn:locallearningcurveempirical3} is similar to \eref{eqn:globallearningcurve}: information from the rest of the graph provides an effective prior variance at any given vertex; the new element here is that this is normalised by the prior variance in the absence of data which must be computed from the auxiliary variables.

It is difficult to asses a priori the quality of the approximation we have made above, but numerical results (see \sref{sec:results}) show that in practice it is very accurate, surprisingly so, even for graph ensembles with a large spread of unnormalised prior variances. On this basis one might speculate whether the approximation could in fact be exact, but we have not been able to find an argument for this and leave it as an open question.

\section{Results}\label{sec:results}
Using \eref{eqn:globalvarianceupdate}, \eref{eqn:globallearningcurve}, \eref{eqn:localupdatenofunc} and
\eref{eqn:locallearningcurveempirical3}, we can now predict the learning curve
for GP regression with a random walk covariance function on random
graphs from ensembles specified by arbitrary fixed degree
distributions. We compare 
our replica-derived predictions for both globally and locally
normalised kernels against a graph ensemble equivalent of the
approximation presented in \cite{Sollich1999a}, and numerically
simulated learning curves on graphs with 500 vertices,
for a range of noise levels. In what follows, we will fix the
hyperparameters $a$ and $p$ to be $2$ and $10$ respectively, but note
that qualitatively similar results are obtained for other values of
these hyperparameters. 

\begin{figure}
\input{global.tex}
\caption{(Left) Learning curves for GP regression with a globally
normalised kernel with $p=10$ and $a=2$ on Erd\H{o}s-R\'enyi random graphs, for a range of
noise levels. Solid lines with filled circles: numerically simulated
learning curves for graphs of size $V=500$, solid lines with triangles: replica predictions
(see Section \ref{sec:global}), dashed lines: The graph equivalent of
\cite{Sollich1999a}. 
(Right) Analogue of left figure for power law random
graphs.\label{fig:global}} 
\end{figure}
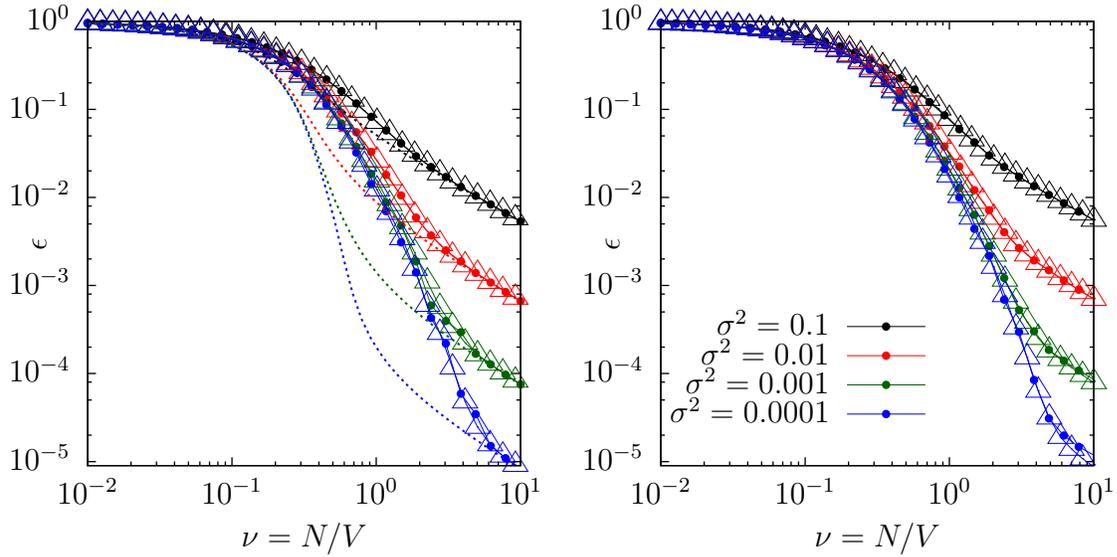

In \fref{fig:global} we compare learning curve predictions for GP
regression with a
globally normalised kernel on two types of random graph
ensembles. \Fref{fig:global} (left) shows results for
the Erd\H{o}s-R\'enyi~\cite{Erdos1959} ensemble. Here each edge in the
graph is present independently with some fixed probability of order
$1/N$. This leads to a Poisson form for the degree distribution, $p_{\lambda}(d)=\lambda^{d}\rme^{-\lambda}/d!$; we
choose the average degree $\lambda$ to be 3. As
can be seen the replica predictions greatly outperform the
approximation from~\cite{Sollich1999a}, especially in the mid-section of the learning curve where fluctuations in the number of
examples at each vertex have the greatest effect (see 
\sref{sec:partfunction} for discussion on why this is to be expected). In fact the replica theory
accurately predicts the numerically simulated learning curve along its
entire length. \Fref{fig:global} (right) similarly assesses our
predictions for a graph ensemble with a power law distribution of
degrees. This generalised random graph ensemble~\cite{Britton2006} is
generated by a superposition of Poisson degree distributions with
different means, $p(d)=\int \rmd\lambda\, p(\lambda) p_{\lambda}(d)$. The distribution of the means $\lambda$ is
set to be a Pareto or power law distribution,
$p(\lambda) =  \alpha\lambda_m^{\alpha}/\lambda^{\alpha+1}$ for
$\lambda\geq\lambda_m$ with the lower cut off $\lambda_m$ set to be 2 
and the exponent $\alpha$ chosen as 2.5. Once more we see that the replica
method faithfully 
predicts the numerically simulated learning curves along their entire
length. Comparison between the graph ensemble equivalent of
\cite{Sollich1999a} (not shown in \fref{fig:global} (right)) and
numerically simulated learning curves again shows that this prediction
fails to accurately predict the mid-section of the learning curve. 

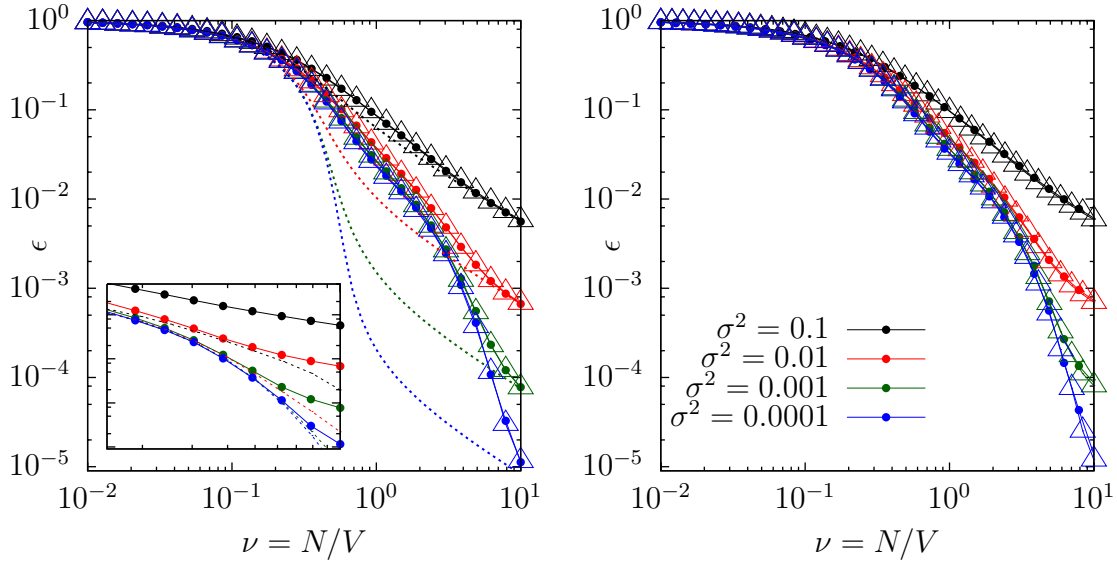
\begin{figure}
\input{local.tex}
\caption{(Left) Learning curves for GP regression with a locally
normalised kernel with $p=10$ and $a=2$ on Erd\H{o}s-R\'enyi random graphs, for a range of
noise levels. Solid lines with 
filled circles: numerically simulated learning curves for graphs of size $V=500$, solid lines with triangles: replica predictions 
(see Section \ref{sec:local}), dashed lines: The graph equivalent of
\cite{Sollich1999a}. 
(Left inset) Dashed lines show the contribution from disconnected
single vertices to the learning curve (solid line with
filled circles).  
(Right) Analogue of left figure for power law random graphs. \label{fig:local}}
\end{figure}

\Fref{fig:local} (left \& right) presents numerically simulated
learning curves and theoretical predictions for a locally
normalised kernel, again on Erd\H{o}s-R\'enyi 
and generalised random graph ensembles with the same parameter
settings as for the globally normalised kernel. Once more
we see that our replica-derived learning curve prediction is
accurate along the whole length of the curve. The graph equivalent of
\cite{Sollich1999a}, however, again fails to accurately predict the
mid-section of the leaning curve. We show in the inset of
\fref{fig:local} (left) a more detailed plot of the somewhat
unexpected shoulder that appears in the learning curves for the
locally normalised kernel. This inset shows that the shoulder is
due to the more realistic normalisation of disconnected
vertices. Around the shoulder one can show that learning curves are
dominated by mean-square errors of predictions on single disconnected
vertices of the graph~\cite{UrryTBA}. In contrast to the globally
normalised case the variance at these vertices is fixed to 1 until an
example is seen. Once on average one 
example has been seen at each disconnected vertex (i.e.\ around
$\nu=1$), the effect of these vertices becomes subdominant and the
learning curves decay in a similar manner as for the globally normalised
kernel. For a more detailed discussion of differences arising from the
choice of global or local 
normalisation for the random walk kernel we refer to~\cite{UrryTBA}. 

\section{Conclusions}\label{sec:conc}
In this paper we focussed on predicting learning curves for GP
regression with random walk kernels of functions defined on large
random graphs. Using the replica method we derived accurate
predictions for two normalisations of the kernel: global
normalisation, where only the average of the prior variance across all
vertices is fixed, and local normalisation, where the prior variance is
uniform across all vertices. The random graph ensembles we considered
are specified by a fixed but arbitrary degree distribution, making the
results applicable to a broad range of graphs. We assume that the
average degree is finite, which implies that the graphs are locally
tree-like. This is, in the end, why the replica approach can give
exact results for this problem in the limit of large graphs. For
simplicity we also assumed that training inputs are drawn from a
uniform distribution across all vertices, though the results
generalise straightforwardly to more general input distributions.

We began in \sref{sec:partfunction} by rewriting
the generalisation error \eref{eqn:epgdef} in terms of the partition
function of a complex-valued graphical
model \eref{eqn:Zsites}.
In \sref{sec:global} we calculated the disorder average of the log
partition function, for the case of a globally normalised
kernel, applying a replica method similar to that of~\cite{Kuhn2007}.

\Sref{sec:local} then dealt with the case of a
locally normalised kernel. Here the local normalisation constants are
determined indirectly as the
local prior variances of the unnormalised GP, and this non-local
dependence on the quenched disorder (i.e.\ the graph structure)
renders the analysis rather more complicated. We used a delta
enforcement term to fix the local normalisers (equation
\eref{eqn:localZdelta}); rewriting this in its
Fourier form and introducing an auxiliary set of replicas we
were once again able to cast the relevant partition function in
the form of a graphical model.
The rest of the analysis then required an additional saddle point method
\emph{inside} the usual saddle point calculation.

Finally, in \sref{sec:results} we compared our new predictions
to an earlier approximation by Sollich~\cite{Sollich1999a},
suitably extended to the graph case, and numerically simulated learning
curves. \Fref{fig:global} for globally
normalised kernels showed that our predictions are accurate
along the whole length of the learning curve for both Erd\H{o}s-R\'{e}nyi and
power law generalised random graph ensembles and for a range of noise
levels. In \fref{fig:local} we showed similar results for the locally
normalised case. We also briefly discussed the change in the shape of
the learning curve that is caused by a change in normalisation from
global to local.

In the future, it would be interesting to extend our approach to a
larger class of graph ensembles, which embody graph structure beyond
degree distributions. One could do this by applying the techniques in
\cite{Rogers2010} where the replica method has been used to calculate
eigenvalue spectra for graphs with topological constraints and
`generalised' degree distributions. The community structures studied in
\cite{Rogers2010} could be further generalised by using the more general community graphs considered in
\cite{Kuhn2011}, where the authors consider an adjacency matrix with block form and a
sparse connectivity pattern between the blocks.

An open problem is whether and how one could apply the methods from
this paper to the significantly more
complicated case of model mismatch, where teacher and student have
different posterior distributions. In this case one would no longer be
able to simplify the generalisation error to the posterior variance as was done in
\sref{sec:partfunction}, so that explicit averages over both teacher
and student posteriors would need to be
performed analytically. Preliminary work has shown that this would entail
the introduction of an additional two local vertex variables and a
second set of replicas. A novel extension of mismatch for the graph
case would be to consider \emph{graph} mismatch between student and teacher,
where the underlying graph structure assumed by the student is a
`corrupted' version of the graph used by the teacher. Finally,
it would be interesting to study how learning curves of real world
data compare to the average case learning curves considered in this
paper, for suitably chosen graph ensembles.  

\clearpage
\appendix
\section{Details of replica analysis}\label{app:details}
\subsection{Graph normaliser}\label{app:graphnormaliser}
In this section we derive the large $V$ asymptotics of the graph
ensemble normaliser $\mathcal{N}$ in \eref{eqn:probgraph}. This is given as a sum over all
possible graphs $G$; i.e.\ over all adjacency matrices with elements $a_{ij}=a_{ji}$
\begin{eqnarray}
   \fl \mathcal{N} &= \sum_{G}\prod_{i<j}\left[\frac{\bar{d}}{V}\delta_{a_{ij},1} +
  \left(1-\frac{\bar{d}}{V}\right)\delta_{a_{ij},0}\right]\prod_{i}\delta_{d_i,\sum_{j}a_{ij}}\\
\fl  &=\int^{2\pi}_{0}\prod_i\frac{\rmd\hd_i}{2\pi}\rme^{-\rmi\hd_id_i}
  \prod_{i<j}\sum_{a_{ij}\in\{0,1\}}\left[\frac{\bar{d}}{V}\delta_{a_{ij},1} +
  \left(1-\frac{\bar{d}}{V}\right)\delta_{a_{ij},0}\right]\rme^{\rmi a_{ij}(\hd_i+\hd_j)}\label{eqn:NlargeV}\\
\fl  &=\int^{2\pi}_{0}\prod_i\frac{\rmd\hd_i}{2\pi}\rme^{-\rmi\hd_id_i}\prod_{i,j}\exp\left(\frac{\bar{d}}{2V}\left(\rme^{\rmi(\hd_i+\hd_j)}
  -1\right)\right),\label{eqn:norm}
\end{eqnarray}
where going from \eref{eqn:NlargeV} to \eref{eqn:norm} we have assumed
that $V$ is large so that $\exp(\frac{x}{V}) \approx 1 + \frac{x}{V}$.
Defining the density $\rho_0 = \frac{1}{V}\sum_i\rme^{\rmi\hd_i}$ with a conjugate enforcement term $\hrho_0$, \eref{eqn:norm}
becomes,
\begin{equation} \label{eqn:normbeforepower}
 \fl \mathcal{N} = \int \rmd\rho_0\rmd\hrho_0\exp\left(V\left[\frac{\bar{d}}{2}(\rho_0^2-1)
  -\rmi\rho_0\hrho_0\right]\right)\prod_i\int^{2\pi}_{0}\frac{\rmd\hd_i}{2\pi}\rme^{-\rmi d_i\hd_i}\rme^{\rmi\hrho_0\sum_{i}\exp(\rmi\hd_i)}.
\end{equation}
Rewriting the final exponential in terms of its power series and performing the integration over $\hd_i$ gives
\begin{equation}\label{eqn:normafterpower}
  \mathcal{N}=\int \rmd\rho_0\rmd\hrho_0\exp\left(V\left[\frac{\bar{d}}{2}\{\rho_0^2-1\}
  -\rmi\rho_0\hrho_0\right]\right)\prod_i\frac{\left(\rmi\hrho_0\right)^{d_i}}{d_i!}.
\end{equation}
Defining the degree distribution
$p(d)=({1}/{V})\sum_i\delta_{d_i,d}$ as usual and bringing the final term into the exponent of the first gives us a form
amenable to saddle point evaluation:
\begin{equation}
 \fl\mathcal{N} =\int \rmd\rho_0\rmd\hrho_0\exp\left(V\left[\frac{\bar{d}}{2}(\rho_0^2-1)
  -\rmi\rho_0\hrho_0 + \sum_{d}p(d)\log\frac{\left(\rmi\hrho_0\right)^{d}}{d!} \right]\right).
\end{equation}
The saddle point of the exponent occurs when
\begin{eqnarray}
  0 &= \bar{d}\rho_0 - \rmi\hrho_0\\
  0 &= -\rmi\rho_0 + \sum_dp(d)d\hrho_0^{-1}
\end{eqnarray}
and solving gives $\rho_0=1$ and $\hrho_0=-\rmi\bar{d}$.

\subsection{Global normalisation}\label{app:global}
We detail in this section some of derivations of the equations in
\sref{sec:global} related to the learning curve for globally
normalised kernels.

\subsubsection{Deriving equation \eref{eqn:globalZpresaddle} from \eref{eqn:Zglobalsites}:}\label{app:globalpresaddle}

Performing the graph average as in \ref{app:graphnormaliser},  \eref{eqn:Zglobalsites} becomes,
\begin{equation}\label{eqn:globalZgraphavg}
  \eqalign{
  \fl
  \langle Z^{n}\rangle_{\x,\mathcal{G}} = \left\langle\int_{0}^{2\pi}\prod_{i}\frac{\rmd\hd_i}{2\pi} \prod_{a=1}^{n}\rmd\hpvecvec^{a}\prod_{i}
  \exp\left(-\sum_{a=1}^{n}\mathcal{H}(\hpvec^a_i,d_i,\kappa,\ni[i])-\rmi\hd_id_i\right)\right.\\
  \left.\times
  \prod_{i,j}\exp\left(\frac{\bar{d}}{2V}\left[\exp\left(-\sum_{a=1}^{n}\mathcal{J}(\hpvec^a_i,\hpvec^a_j)+\rmi(\hd_i+\hd_j)\right)-1\right]\right)\right\rangle_{\x}.}
\end{equation}
Using the replica densities defined in \eref{eqn:globaldensities} we see that the final product in \eref{eqn:globalZgraphavg}
can be rewritten in terms of \eref{eqn:globalS1} by moving the product over $i,j$ into the exponent and realising
that summing over all $\hpvec_i^a$ and $\hpvec_j^a$ is the same as integrating over the replica densities. This gives the
term $\exp(\frac{V\bar{d}}{2}(S_{1}-1))$ in \eref{eqn:globalZpresaddle}.

In order to use \eref{eqn:globaldensities} in \eref{eqn:Zglobalsites} we must introduce a delta enforcement term. $S_2$, defined in
\eref{eqn:globalS2}, is a direct consequence of including the functional delta term,
\begin{equation}\label{eqn:globalreplicaenforcement}
  \eqalign{
  \fl\delta\left(V\rhofuncshort - \sum_i\prod_{a=1}^{n}\delta(\hpvec^a-\hpvec^a_i)\rme^{\rmi\hd_i}
\right) = \\ \int \mathcal{D}\hrho \exp\left(\rmi\int \prod_a\rmd\hpvec^a \hrhofuncshort\left(\sum_i\prod_{a=1}^{n}\delta(\hpvec^a-\hpvec^a_i)\rme^{\rmi\hd_i}-V\rhofuncshort 
  \right)\right).}
\end{equation}
Finally $S_{3}$, defined in \eref{eqn:globalS3}, is derived from the first product in \eref{eqn:globalZgraphavg} with
the remaining term from \eref{eqn:globalreplicaenforcement}. These
factors give
\begin{equation}
  \eqalign{
\fl\left\langle\int_{0}^{2\pi}\frac{\rmd\bhd}{(2\pi)^V} \prod_{a=1}^{n}\rmd\hpvecvec^{a}\prod_{i}
  \exp\left(-\sum_{a=1}^{n}\mathcal{H}(\hpvec^a_i,d_i,\kappa,\ni[i])-\rmi\hd_id_i +
 \rmi\hrhofuncsiteshort\rme^{\rmi\hd_i}\right)\right\rangle_{\x},}
\end{equation}
where $\hrhofuncsiteshort = \hrhofuncsite$. Bringing the integration
over $\hpvecvec$ into the exponent by using a logarithm and performing
the same procedure as in \ref{app:graphnormaliser}, in going from
\eref{eqn:normbeforepower} to \eref{eqn:normafterpower}, gives
\begin{equation}
  \eqalign{
\fl\exp\left(
\log \left\langle\int \prod_{a=1}^{n}\rmd\hpvecvec^{a}\prod_{i}
  \exp\left(-\sum_{a=1}^{n}\mathcal{H}(\hpvec^a_i,d_i,\kappa,\ni[i])\right)\frac{(\rmi\hrhofuncsiteshort)^{d_i}}{d_i!}\right\rangle_{\x}\right).}
\end{equation}
Since we integrate over each site independently and also the data average separates into independent Poisson averages over the number of examples $\ni[i]$ at each vertex, we can introduce \mbox{$p(d) =
1/V\sum_{i}\delta_{d_i,d}$} and replace the integral over $\hpvec_i^a$
with an integral over $\hpvec^a$. This yields the term $\exp(VS_{3})$
in \eref{eqn:globalZpresaddle}.

\subsubsection{Deriving equation \eref{eqn:globalupdate} from \eref{eqn:globalZpresaddle}:}\label{app:globalupdate}
Substituting \eref{eqn:globalansatzrho} \& \eref{eqn:globalansatzhrho}
into \eref{eqn:globalS1} to \eref{eqn:globalS3} we have 
\begin{eqnarray}
  \eqalign{
 \fl S_1[\rho] = \int \mathcal{D}\psi\mathcal{D}\psi'\pi[\psi]\pi[\psi']\int \prod_{a=1}^{n}\rmd\hpvec^a
 \rmd(\hpvec^a)'\\\prod_{a=1}^{n}\frac{\exp\left(-\psifunc[a]-\psifuncdash[a]\right)}{Z[\psi]Z[\psi']}
 \exp\left(-\mathcal{J}(\hpvec^a,(\hpvec^a)')\right)\label{eqn:globalansatzS1}}\\
 \fl S_{2}[\rho,\hrho] =-\rmi\bar{d}\int
\mathcal{D}\psi\mathcal{D}\hpsi\,
\pi[\psi]\hpi[\hpsi] \int\prod_{a=1}^{n}\rmd\hpvec^a
 \prod_{a=1}^{n}\frac{\exp\left(-\psifunc[a]-\hpsifunc[a]\right)}{Z[\psi]Z[\hpsi]}\label{eqn:globalansatzS2}\\
 \eqalign{
 \fl S_{3}[\hrho] = \sum_{d}p(d)\log\left\langle\frac{\bar{d}^{d}}{d!}\int
 \prod_{i=1}^{d}\mathcal{D}\hpsi^{i}\hpi[\hpsi^{i}]\int\prod_{a=1}^{n}\rmd\hpvec^a\right.\\\left.\exp\left(-\sum_{a=1}^{n}\mathcal{H}(\hpvec^a,d,\kappa,\ni)\right)
 \prod_{i=1}^{d}\prod_{a=1}^{n}\frac{\exp\left(-\hpsifunc[a][i]\right)}{Z[\hpsi^i]}\right\rangle_\gamma.}\label{eqn:globalansatzS3}
\end{eqnarray}
We now expand to order $n$, using $x^n = 1+ n\log(x) + \ldots$
This gives
\begin{eqnarray}
  \eqalign{
 \fl S_1[\rho] =\int \mathcal{D}\psi\mathcal{D}\psi'\pi[\psi]\pi[\psi']\left[\rule{0cm}{0.9cm}\right. 1 +\\ n\log\int\rmd\hpvec
 \rmd\hpvec'\frac{\exp\left(-\psifunc-\psifuncdash-\mathcal{J}(\hpvec,\hpvec')\right)}{Z[\psi]Z[\psi']}
 \left.\rule{0cm}{0.9cm}\right]\label{eqn:globallogxS1}}\\
 \fl S_{2}[\rho,\hrho] =-i\bar{d}\int \mathcal{D}\psi\mathcal{D}\hpsi \,\pi[\psi]\hpi[\hpsi] \left[1+n\log\int\rmd\hpvec
 \frac{\exp\left(-\psifunc-\hpsifunc[][]\right)}{Z[\psi]Z[\hpsi]}\right]\label{eqn:globallogxS2}\\
 \eqalign{
 \fl S_{3}[\hrho] = \sum_{d}p(d)\log\left[\frac{\bar{d}^{d}\int\prod_{i=1}^{d}\mathcal{D}\hpsi^{i}\hpi[\hpsi^{i}]}{d!}\right] \\
 +n\sum_d p(d)\int
 \prod_{i=1}^{d}\mathcal{D}\hpsi^{i}\hpi[\hpsi^{i}]\left\langle
\log\int\rmd\hpvec\exp\left(-\mathcal{H}(\hpvec,d,\kappa,\ni)\right)\right.\\\left.
\times
 \prod_{i=1}^{d}\frac{\exp\left(-\hpsifunc[][i]\right)}{Z[\hpsi^{i}]}\right\rangle_\gamma
\left[\int\prod_{i=1}^{d}\mathcal{D}\hpsi^{i}\hpi[\hpsi^{i}]\right]^{-1}.}\label{eqn:globallogxS3}
\end{eqnarray}

The leading $O(1)$ contributions yield as saddle point conditions for
$\pi$ and $\hpi$ simply $\int\mathcal{D}\psi\,\pi[\psi]=1$ and
$\int\mathcal{D}\hpsi\,\hpi[\hpsi]=1$. With these normalisation
constraints inserted, the $O(1)$ terms take the same form as in \ref{app:graphnormaliser}. Thus for $V\to\infty$ the
$O(1)$ terms in the exponent of \eref{eqn:globalZpresaddle} will
cancel, and we are left as expected with the $\lambda$-dependent
subleading $O(n)$ terms. These lead to the saddle point equations for $\pi[\psi]$ and $\hpi[\hpsi]$
\begin{eqnarray}
  \eqalign{
\label{eq:simpler_SP}
\fl \mu = \bar{d}\int \mathcal{D}\psi' \pi[\psi']
\log 
\int\rmd\hpvec\rmd\hpvec'\frac{\exp\left(-\psifunc-\psifuncdash-\mathcal{J}(\hpvec,\hpvec')\right)}{Z[\psi]Z[\psi']}
\\
-\bar{d}\int\mathcal{D}\hpsi\,
\hpi[\hpsi]\log\int\rmd\hpvec\frac{\exp\left(-\psifunc-\hpsifunc[][]\right)}{Z[\psi]Z[\hpsi]}}\\
\eqalign{
\label{eq:second_SP}
\fl\hmu =
\sum_dp(d)d\int
\prod_{i=1}^{d-1}\mathcal{D}\hpsi^{i}\hpi[\hpsi^{i}]\\
\times\left\langle\log\int\rmd\hpvec\frac{\exp\left(-\mathcal{H}(\hpvec,d,\kappa,\ni)-\hpsifunc[][]-\sum_{i=1}^{d-1}\hpsifunc[][i]\right)
}{Z[\hpsi]\prod_{i=1}^{d}Z[\hpsi^i]}\right\rangle_{\gamma}\\
-\bar{d}\int\mathcal{D}\psi\,\pi[\psi]\log\int\rmd\hpvec\frac{\exp\left(-\psifunc-\hpsifunc[][]\right)}{Z[\psi]Z[\hpsi]},}
\end{eqnarray}
 where the $O(1)$ constraints have been enforced with Lagrange multipliers 
$\mu$ and $\hmu$.
%
Looking at \eref{eq:simpler_SP}, the $Z[\psi']$ and $Z[\hpsi]$ factors
give additive constants that can be absorbed, along with the factors
of $\bar{d}$, into a redefinition of $\mu$. Bearing in mind the
definition \eref{eqn:globalhPsi} of $\hPsi$, the resulting equation
has the form
\begin{equation}
\mu = \int \mathcal{D}\psi' \pi[\psi']
F(\psi,\hPsi[\psi'])  
-\int\mathcal{D}\hpsi\,\hpi[\hpsi] F(\psi,\hpsi)
\end{equation}
in which $F(\psi,\hpsi) =
\log
\int\rmd\hpvec \,Z^{-1}[\psi]\exp\left(-\psifunc-\hpsifunc[][]\right)
$. As this equation must hold for arbitrary $\psi$, it follows
that the distribution of $\hpsi$ must equal the distribution of
$\hPsi[\psi']$, hence
\begin{equation}
  \hpi[\hpsi] = \int \mathcal{D}\psi\pi[\psi]\delta(\hpsi -
\hPsi[\psi]).
\label{eq:simpler_SP_final}
\end{equation}
(To be precise, $\hpsi$ and $\hPsi[\psi']$ can differ by an arbitrary
additive constant because $F(\psi,\hpsi+\mbox{const}) = 
F(\psi,\hpsi)+\mbox{const}$; we fix this constant to the most
convenient value in writing \eref{eq:simpler_SP_final}.) Similarly one derives 
from \eref{eq:second_SP} that
\begin{equation}
  \pi[\psi] = \sum_{d}\frac{p(d)d}{\bar{d}}\left\langle\int
  \prod_{i=1}^{d-1}\mathcal{D}\hpsi^{i}\hpi[\hpsi^{i}]\delta(\psi -
  \Psi[\hpsi^{1},\ldots,\hpsi^{d-1}])\right\rangle_{\gamma},
\end{equation}
with $\Psi$ defined in \eref{eqn:globalPsi}.
Combining this with \eref{eq:simpler_SP_final} gives rise to
\eref{eqn:globalupdate}.

\subsubsection{Woodbury derivations} \label{app:globalwoodburyderivations}
The apparent singularity in the update equations given by \eref{eqn:globalvarianceupdate} can be eliminated by using the
Woodbury identity for inverting matrices~\cite{Hager1989}. We may
write
\begin{equation}\label{eqn:woodbury}
 \fl\left(\e0\frac{d\kappa}{\ni/\sigma^{2}+\lambda}\e0\T + \M_{d-1}\right)^{-1} = \M_{d-1}^{-1} -
  \frac{\M_{d-1}^{-1}\e0\e0\T\M_{d-1}^{-1}}{\left(\ni/\sigma^{2}+\lambda\right)/(d\kappa) + \e0\T\M_{d-1}^{-1}\e0},
\end{equation}
where $\M_{d-1} = \O - \sum_{i=1}^{d-1}\X\V^{i}\X -
\e0\frac{d\kappa}{\ni/\sigma^{2}+\lambda}\e0\T$ contains only
$\lambda$-independent terms. On the right hand side of \eref{eqn:woodbury} we
can now take $\lambda\to 0$ without problems, even when $\gamma=0$.

A similar method may be
applied to arrive at \eref{eqn:globallearningcurve}. Direct differentiation with respect to $\lambda$ of
$S_{3}^{O(n)}$, defined as the coefficient multiplying $n$ in \eref{eqn:globallogxS3} will result in the following expression for the
generalisation error: 
\begin{equation}\label{eqn:globalsingularlearningcurve}
\fl  \epsilon_{g} = \lim_{\lambda\to0}\sum_{d}p(d)\left\langle\frac{1}{\ni/\sigma^{2}+\lambda}\left[1 -
  \frac{d\kappa}{\ni/\sigma^{2}+\lambda}\e0\T(\O - \sum_{i=1}^{d}\X\V^{i}\X)^{-1}\e0\right]\right\rangle_{\gamma}.
\end{equation}
If we now define $\M_{d}$ in analogy to $\M_{d-1}$ above, but with the sum
over $i$ running from 1 to $d$, we can substitute
the analogue of \eref{eqn:woodbury} into
\eref{eqn:globalsingularlearningcurve} and 
take $\lambda\to0$ to arrive at \eref{eqn:globallearningcurve}. 

\subsection{Local normalisation}\label{app:local}
We detail in this section some of the derivations of equations in \sref{sec:local} related to the locally normalised kernel case.
\subsubsection{Deriving equation \eref{eqn:locallargeLPhi} from equations \eref{eqn:localPsi} and \eref{eqn:localhPsi}:}\label{app:localO(n)derive}

In the main text we discussed that for the calculation of $\hPsi[\psi]$ (see \eref{eqn:localhPsi}), only the (logarithmic) marginal $\phi$ of the input function $\psi$ over $\kappa$ and $\hkappa$ is required. The corresponding marginal of the functional $\Psi[\hpsi^{1},\ldots,\hpsi^{d-1}]$ in \eref{eqn:localPsi} is then
%
\begin{equation}\label{eqn:internalsaddle}
  \eqalign{ 
\fl
\Phi[\hphi^{1},\ldots,\hphi^{d-1}] = -\log
\int\frac{\rmd\kappa\rmd\hkappa L}{2\pi }\zeta(\hpvec,\kappa,\hkappa,\{\hpvec\aux^{l}\})\exp\left(L\xi(\{\hpvec\aux^{l}\},\kappa,\hkappa)\right)},
\end{equation}
with
\begin{eqnarray}
    \zeta(\hpvec,\kappa,\hkappa,\{\hpvec\aux^{l}\}) =
    \exp\left(-\mathcal{H}(\hpvec,d,\kappa,\gamma)-\sum_{i=1}^{d}\hphi^{i}(\hpvec,\{\hpvec\aux^{l}\})\right)\label{eqn:zetafunc}\\
  \xi(\{\hpvec\aux^{l}\},\kappa,\hkappa)=-\frac{1}{L}\sum_l\mathcal{H}\aux(\hpvec\aux^l,d,\hkappa) +\rmi\kappa\hkappa.
\end{eqnarray}
As in the main text we have written $\hphi^i$ here instead of $\hpsi^i$ to emphasize that these functions are independent of $\kappa$ and $\hkappa$.

For large $L$ the integral in \eref{eqn:internalsaddle} over $\kappa$ and $\hkappa$ will be dominated by its saddle point. The saddle point conditions are
\begin{eqnarray}
  0 = \rmi\hkappa\\
  0 =  \left[-\frac{\rmi}{\lambda\aux+2\rmi\hkappa} + \rmi\kappa +
  \rmi \frac{d}{L}\sum_l \frac{\e0\T \hpvec^{l}\aux(\hpvec^{l}\aux)\T\e0}{(\lambda\aux+2\rmi\hkappa)^2}\right].
\end{eqnarray}
This gives us the saddle point values
\begin{eqnarray}\label{eqn:kappasol}
  \kappa = \frac{1}{\lambda\aux}-\frac{d}{L}\sum_l \frac{\e0\T \hpvec^{l}\aux(\hpvec^{l}\aux)\T\e0}{\lambda\aux^2} = v\Big(\frac{1}{L}\sum_{l}\hpvec\aux^{l}(\hpvec\aux^{l})\T\Big)\\
  \hkappa = 0.
\end{eqnarray}
One subtlety here is that we need to make sure that the integral over
$\kappa$ and $\hkappa$ in \eref{eqn:internalsaddle} is evaluated accurately 
including $O(1)$ prefactors, because such contributions will affect the $\hpvec$-dependence of $\hPhi$. To obtain these prefactors one needs to
expand the exponent of \eref{eqn:internalsaddle} to second order
about the saddle point. Fortunately the relevant Hessian of 
$\xi(\{\hpvec\aux^l\},\kappa,\hkappa)$
has a constant determinant, and the factor from the fluctuation
correction is simply $(2\pi/L)$. This just cancels the scaling factor
in front of the integral over $\kappa$ and $\hkappa$. Substituting the saddle point approximation
into $\Phi[\hphi^{1},\ldots,\hphi^{d-1}]$ then gives \eref{eqn:locallargeLPhi} in the main text.
\section*{References}
\bibliographystyle{unsrt}
\bibliography{refs}
\end{document}

%% file: population.tex
\begingroup
  \makeatletter
  \providecommand\color[2][]{%
    \GenericError{(gnuplot) \space\space\space\@spaces}{%
      Package color not loaded in conjunction with
      terminal option `colourtext'%
    }{See the gnuplot documentation for explanation.%
    }{Either use 'blacktext' in gnuplot or load the package
      color.sty in LaTeX.}%
    \renewcommand\color[2][]{}%
  }%
  \providecommand\includegraphics[2][]{%
    \GenericError{(gnuplot) \space\space\space\@spaces}{%
      Package graphicx or graphics not loaded%
    }{See the gnuplot documentation for explanation.%
    }{The gnuplot epslatex terminal needs graphicx.sty or graphics.sty.}%
    \renewcommand\includegraphics[2][]{}%
  }%
  \providecommand\rotatebox[2]{#2}%
  \@ifundefined{ifGPcolor}{%
    \newif\ifGPcolor
    \GPcolortrue
  }{}%
  \@ifundefined{ifGPblacktext}{%
    \newif\ifGPblacktext
    \GPblacktexttrue
  }{}%
  \let\gplgaddtomacro\g@addto@macro
  \gdef\gplbacktext{}%
  \gdef\gplfronttext{}%
  \makeatother
  \ifGPblacktext
    \def\colorrgb#1{}%
    \def\colorgray#1{}%
  \else
    \ifGPcolor
      \def\colorrgb#1{\color[rgb]{#1}}%
      \def\colorgray#1{\color[gray]{#1}}%
      \expandafter\def\csname LTw\endcsname{\color{white}}%
      \expandafter\def\csname LTb\endcsname{\color{black}}%
      \expandafter\def\csname LTa\endcsname{\color{black}}%
      \expandafter\def\csname LT0\endcsname{\color[rgb]{1,0,0}}%
      \expandafter\def\csname LT1\endcsname{\color[rgb]{0,1,0}}%
      \expandafter\def\csname LT2\endcsname{\color[rgb]{0,0,1}}%
      \expandafter\def\csname LT3\endcsname{\color[rgb]{1,0,1}}%
      \expandafter\def\csname LT4\endcsname{\color[rgb]{0,1,1}}%
      \expandafter\def\csname LT5\endcsname{\color[rgb]{1,1,0}}%
      \expandafter\def\csname LT6\endcsname{\color[rgb]{0,0,0}}%
      \expandafter\def\csname LT7\endcsname{\color[rgb]{1,0.3,0}}%
      \expandafter\def\csname LT8\endcsname{\color[rgb]{0.5,0.5,0.5}}%
    \else
      \def\colorrgb#1{\color{black}}%
      \def\colorgray#1{\color[gray]{#1}}%
      \expandafter\def\csname LTw\endcsname{\color{white}}%
      \expandafter\def\csname LTb\endcsname{\color{black}}%
      \expandafter\def\csname LTa\endcsname{\color{black}}%
      \expandafter\def\csname LT0\endcsname{\color{black}}%
      \expandafter\def\csname LT1\endcsname{\color{black}}%
      \expandafter\def\csname LT2\endcsname{\color{black}}%
      \expandafter\def\csname LT3\endcsname{\color{black}}%
      \expandafter\def\csname LT4\endcsname{\color{black}}%
      \expandafter\def\csname LT5\endcsname{\color{black}}%
      \expandafter\def\csname LT6\endcsname{\color{black}}%
      \expandafter\def\csname LT7\endcsname{\color{black}}%
      \expandafter\def\csname LT8\endcsname{\color{black}}%
    \fi
  \fi
  \setlength{\unitlength}{0.0500bp}%
  \begin{picture}(8616.00,4308.00)%

    \gplgaddtomacro\gplbacktext{%
      \csname LTb\endcsname%
      \put(528,704){\makebox(0,0)[r]{\strut{}$10^{-3}$}}%
      \put(528,1372){\makebox(0,0)[r]{\strut{}$10^{-2}$}}%
      \put(528,2040){\makebox(0,0)[r]{\strut{}$10^{-1}$}}%
      \put(528,2707){\makebox(0,0)[r]{\strut{}$10^{0}$}}%
      \put(528,3375){\makebox(0,0)[r]{\strut{}$10^{1}$}}%
      \put(528,4043){\makebox(0,0)[r]{\strut{}$10^{2}$}}%
      \put(795,484){\makebox(0,0){\strut{} 0}}%
      \put(1333,484){\makebox(0,0){\strut{} 0.2}}%
      \put(1872,484){\makebox(0,0){\strut{} 0.4}}%
      \put(22,3473){\rotatebox{-270}{\makebox(0,0){\strut{}$\log p(V_{00})$}}}%
      \put(1535,154){\makebox(0,0){\strut{}$V_{00}$}}%
    }%
    \gplgaddtomacro\gplfronttext{%
    }%
    \gplgaddtomacro\gplbacktext{%
      \csname LTb\endcsname%
      \put(2549,484){\makebox(0,0){\strut{} 0}}%
      \put(3091,484){\makebox(0,0){\strut{} 0.2}}%
      \put(3633,484){\makebox(0,0){\strut{} 0.4}}%
      \put(4175,484){\makebox(0,0){\strut{} 0.6}}%
      \put(4307,704){\makebox(0,0)[l]{\strut{}}}%
      \put(4307,1372){\makebox(0,0)[l]{\strut{}}}%
      \put(4307,2040){\makebox(0,0)[l]{\strut{}}}%
      \put(4307,2707){\makebox(0,0)[l]{\strut{}}}%
      \put(4307,3375){\makebox(0,0)[l]{\strut{}}}%
      \put(4307,4043){\makebox(0,0)[l]{\strut{}}}%
      \put(3294,154){\makebox(0,0){\strut{}$V_{00}$}}%
    }%
    \gplgaddtomacro\gplfronttext{%
    }%
    \gplgaddtomacro\gplbacktext{%
      \csname LTb\endcsname%
      \put(4836,704){\makebox(0,0)[r]{\strut{}$10^{-3}$}}%
      \put(4836,1372){\makebox(0,0)[r]{\strut{}$10^{-2}$}}%
      \put(4836,2040){\makebox(0,0)[r]{\strut{}$10^{-1}$}}%
      \put(4836,2707){\makebox(0,0)[r]{\strut{}$10^{0}$}}%
      \put(4836,3375){\makebox(0,0)[r]{\strut{}$10^{1}$}}%
      \put(4836,4043){\makebox(0,0)[r]{\strut{}$10^{2}$}}%
      \put(5073,484){\makebox(0,0){\strut{} 0}}%
      \put(5492,484){\makebox(0,0){\strut{} 0.2}}%
      \put(5912,484){\makebox(0,0){\strut{} 0.4}}%
      \put(6331,484){\makebox(0,0){\strut{} 0.6}}%
      \put(6751,484){\makebox(0,0){\strut{} 0.8}}%
      \put(7170,484){\makebox(0,0){\strut{} 1}}%
      \put(7590,484){\makebox(0,0){\strut{} 1.2}}%
      \put(8009,484){\makebox(0,0){\strut{} 1.4}}%
      \put(4330,3473){\rotatebox{-270}{\makebox(0,0){\strut{}$\log p(V_{00})$}}}%
      \put(6593,154){\makebox(0,0){\strut{}$V_{00}$}}%
    }%
    \gplgaddtomacro\gplfronttext{%
      \csname LTb\endcsname%
      \put(7232,3870){\makebox(0,0)[r]{\strut{}$\nu=10^{-4}$}}%
      \csname LTb\endcsname%
      \put(7232,3650){\makebox(0,0)[r]{\strut{}$\nu=10^{0}$}}%
      \csname LTb\endcsname%
      \put(7232,3430){\makebox(0,0)[r]{\strut{}$\nu=10^{1}$}}%
    }%
    \gplbacktext
    \put(0,0){\includegraphics{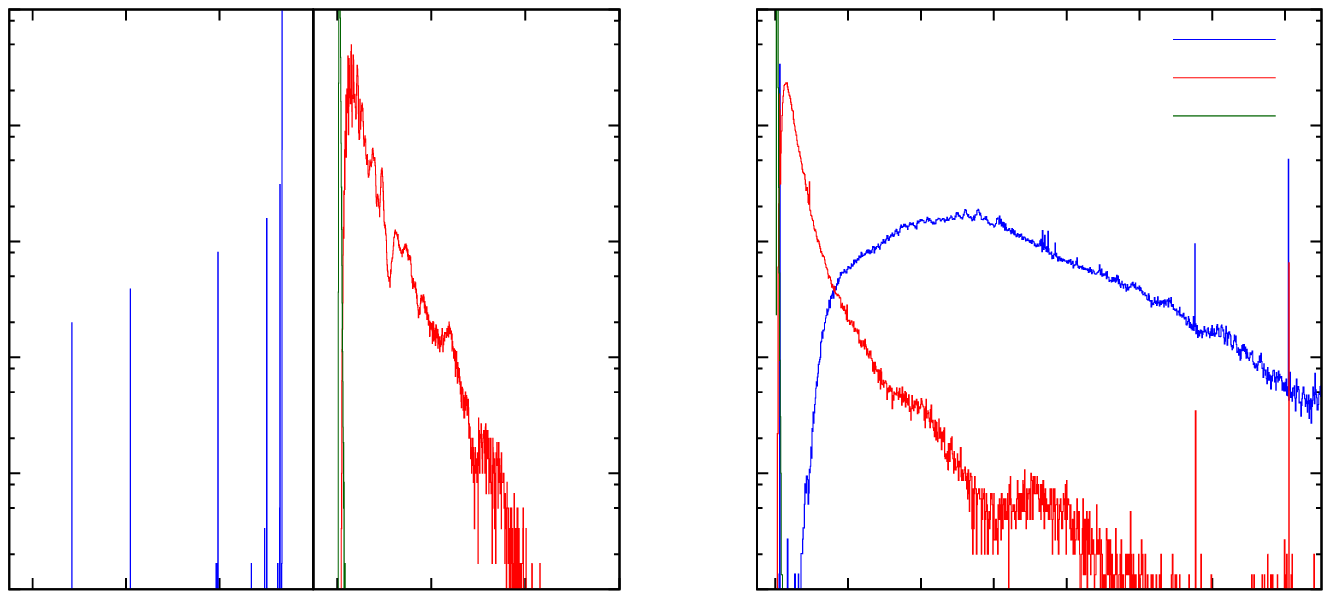}}%
    \gplfronttext
     \put(1200,3400){\makebox(0,0){(a)}}%
     \put(3000,3400){\makebox(0,0){(b)}}%
     \put(5600,3400){\makebox(0,0){(c)}}%
  \end{picture}%
\endgroup

%% file: global.tex
\begingroup
  \makeatletter
  \providecommand\color[2][]{%
    \GenericError{(gnuplot) \space\space\space\@spaces}{%
      Package color not loaded in conjunction with
      terminal option `colourtext'%
    }{See the gnuplot documentation for explanation.%
    }{Either use 'blacktext' in gnuplot or load the package
      color.sty in LaTeX.}%
    \renewcommand\color[2][]{}%
  }%
  \providecommand\includegraphics[2][]{%
    \GenericError{(gnuplot) \space\space\space\@spaces}{%
      Package graphicx or graphics not loaded%
    }{See the gnuplot documentation for explanation.%
    }{The gnuplot epslatex terminal needs graphicx.sty or graphics.sty.}%
    \renewcommand\includegraphics[2][]{}%
  }%
  \providecommand\rotatebox[2]{#2}%
  \@ifundefined{ifGPcolor}{%
    \newif\ifGPcolor
    \GPcolortrue
  }{}%
  \@ifundefined{ifGPblacktext}{%
    \newif\ifGPblacktext
    \GPblacktexttrue
  }{}%
  \let\gplgaddtomacro\g@addto@macro
  \gdef\gplbacktext{}%
  \gdef\gplfronttext{}%
  \makeatother
  \ifGPblacktext
    \def\colorrgb#1{}%
    \def\colorgray#1{}%
  \else
    \ifGPcolor
      \def\colorrgb#1{\color[rgb]{#1}}%
      \def\colorgray#1{\color[gray]{#1}}%
      \expandafter\def\csname LTw\endcsname{\color{white}}%
      \expandafter\def\csname LTb\endcsname{\color{black}}%
      \expandafter\def\csname LTa\endcsname{\color{black}}%
      \expandafter\def\csname LT0\endcsname{\color[rgb]{1,0,0}}%
      \expandafter\def\csname LT1\endcsname{\color[rgb]{0,1,0}}%
      \expandafter\def\csname LT2\endcsname{\color[rgb]{0,0,1}}%
      \expandafter\def\csname LT3\endcsname{\color[rgb]{1,0,1}}%
      \expandafter\def\csname LT4\endcsname{\color[rgb]{0,1,1}}%
      \expandafter\def\csname LT5\endcsname{\color[rgb]{1,1,0}}%
      \expandafter\def\csname LT6\endcsname{\color[rgb]{0,0,0}}%
      \expandafter\def\csname LT7\endcsname{\color[rgb]{1,0.3,0}}%
      \expandafter\def\csname LT8\endcsname{\color[rgb]{0.5,0.5,0.5}}%
    \else
      \def\colorrgb#1{\color{black}}%
      \def\colorgray#1{\color[gray]{#1}}%
      \expandafter\def\csname LTw\endcsname{\color{white}}%
      \expandafter\def\csname LTb\endcsname{\color{black}}%
      \expandafter\def\csname LTa\endcsname{\color{black}}%
      \expandafter\def\csname LT0\endcsname{\color{black}}%
      \expandafter\def\csname LT1\endcsname{\color{black}}%
      \expandafter\def\csname LT2\endcsname{\color{black}}%
      \expandafter\def\csname LT3\endcsname{\color{black}}%
      \expandafter\def\csname LT4\endcsname{\color{black}}%
      \expandafter\def\csname LT5\endcsname{\color{black}}%
      \expandafter\def\csname LT6\endcsname{\color{black}}%
      \expandafter\def\csname LT7\endcsname{\color{black}}%
      \expandafter\def\csname LT8\endcsname{\color{black}}%
    \fi
  \fi
  \setlength{\unitlength}{0.0500bp}%
  \begin{picture}(8640.00,4320.00)%
    \gplgaddtomacro\gplbacktext{%
      \csname LTb\endcsname%
      \put(528,734){\makebox(0,0)[r]{\strut{}$10^{-5}$}}%
      \put(528,1399){\makebox(0,0)[r]{\strut{}$10^{-4}$}}%
      \put(528,2063){\makebox(0,0)[r]{\strut{}$10^{-3}$}}%
      \put(528,2727){\makebox(0,0)[r]{\strut{}$10^{-2}$}}%
      \put(528,3391){\makebox(0,0)[r]{\strut{}$10^{-1}$}}%
      \put(528,4055){\makebox(0,0)[r]{\strut{}$10^{0}$}}%
      \put(660,484){\makebox(0,0){\strut{}$10^{-2}$}}%
      \put(1748,484){\makebox(0,0){\strut{}$10^{-1}$}}%
      \put(2835,484){\makebox(0,0){\strut{}$10^{0}$}}%
      \put(3923,484){\makebox(0,0){\strut{}$10^{1}$}}%
      \put(286,2379){\rotatebox{-270}{\makebox(0,0){\strut{}$\epsilon$}}}%
      \put(2291,154){\makebox(0,0){\strut{}$\nu=N/V$}}%
    }%
    \gplgaddtomacro\gplfronttext{%
    }%
    \gplgaddtomacro\gplbacktext{%
      \csname LTb\endcsname%
      \put(4848,734){\makebox(0,0)[r]{\strut{}$10^{-5}$}}%
      \put(4848,1399){\makebox(0,0)[r]{\strut{}$10^{-4}$}}%
      \put(4848,2063){\makebox(0,0)[r]{\strut{}$10^{-3}$}}%
      \put(4848,2727){\makebox(0,0)[r]{\strut{}$10^{-2}$}}%
      \put(4848,3391){\makebox(0,0)[r]{\strut{}$10^{-1}$}}%
      \put(4848,4055){\makebox(0,0)[r]{\strut{}$10^{0}$}}%
      \put(4980,484){\makebox(0,0){\strut{}$10^{-2}$}}%
      \put(6068,484){\makebox(0,0){\strut{}$10^{-1}$}}%
      \put(7155,484){\makebox(0,0){\strut{}$10^{0}$}}%
      \put(8243,484){\makebox(0,0){\strut{}$10^{1}$}}%
      \put(4606,2379){\rotatebox{-270}{\makebox(0,0){\strut{}$\epsilon$}}}%
      \put(6611,154){\makebox(0,0){\strut{}$\nu=N/V$}}%
    }%
    \gplgaddtomacro\gplfronttext{%
      \csname LTb\endcsname%
      \put(6251,1753){\makebox(0,0)[r]{\strut{}$\sigma^{2} = 0.1$}}%
      \csname LTb\endcsname%
      \put(6251,1533){\makebox(0,0)[r]{\strut{}$\sigma^{2} = 0.01$}}%
      \csname LTb\endcsname%
      \put(6251,1313){\makebox(0,0)[r]{\strut{}$\sigma^{2} = 0.001$}}%
      \csname LTb\endcsname%
      \put(6251,1093){\makebox(0,0)[r]{\strut{}$\sigma^{2} = 0.0001$}}%
    }%
    \gplbacktext
    \put(0,0){\includegraphics{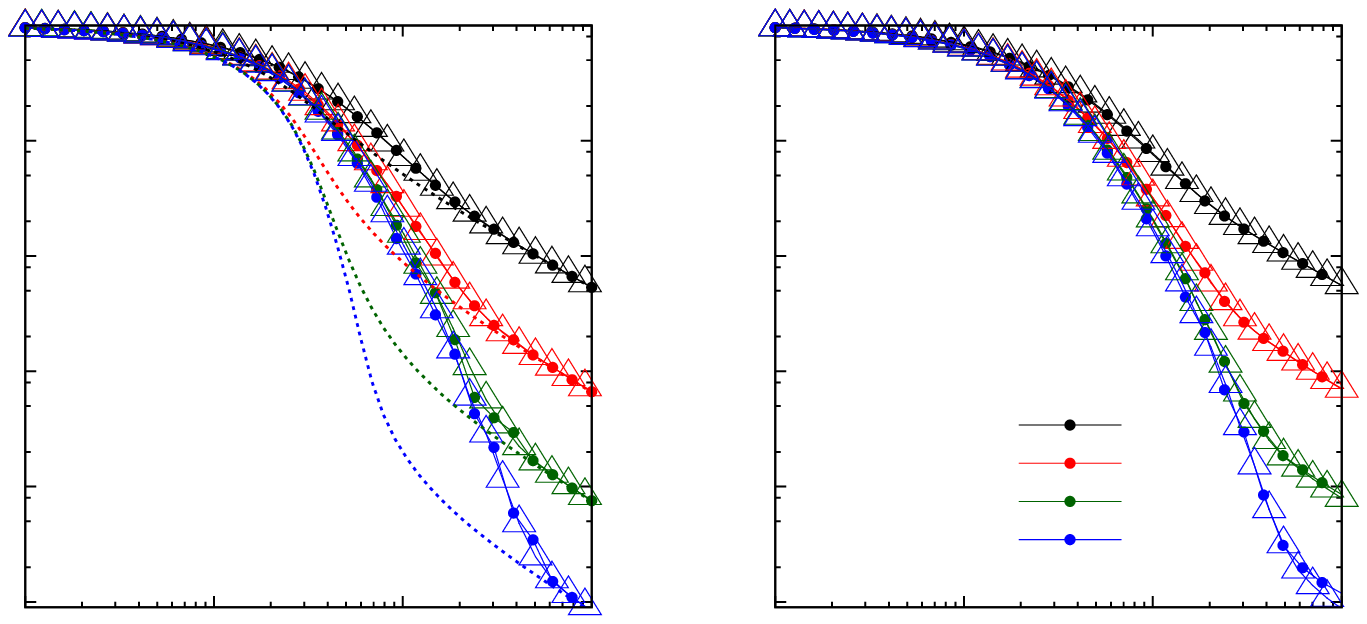}}%
    \gplfronttext
  \end{picture}%
\endgroup

%% file: local.tex
\begingroup
  \makeatletter
  \providecommand\color[2][]{%
    \GenericError{(gnuplot) \space\space\space\@spaces}{%
      Package color not loaded in conjunction with
      terminal option `colourtext'%
    }{See the gnuplot documentation for explanation.%
    }{Either use 'blacktext' in gnuplot or load the package
      color.sty in LaTeX.}%
    \renewcommand\color[2][]{}%
  }%
  \providecommand\includegraphics[2][]{%
    \GenericError{(gnuplot) \space\space\space\@spaces}{%
      Package graphicx or graphics not loaded%
    }{See the gnuplot documentation for explanation.%
    }{The gnuplot epslatex terminal needs graphicx.sty or graphics.sty.}%
    \renewcommand\includegraphics[2][]{}%
  }%
  \providecommand\rotatebox[2]{#2}%
  \@ifundefined{ifGPcolor}{%
    \newif\ifGPcolor
    \GPcolortrue
  }{}%
  \@ifundefined{ifGPblacktext}{%
    \newif\ifGPblacktext
    \GPblacktexttrue
  }{}%
  \let\gplgaddtomacro\g@addto@macro
  \gdef\gplbacktext{}%
  \gdef\gplfronttext{}%
  \makeatother
  \ifGPblacktext
    \def\colorrgb#1{}%
    \def\colorgray#1{}%
  \else
    \ifGPcolor
      \def\colorrgb#1{\color[rgb]{#1}}%
      \def\colorgray#1{\color[gray]{#1}}%
      \expandafter\def\csname LTw\endcsname{\color{white}}%
      \expandafter\def\csname LTb\endcsname{\color{black}}%
      \expandafter\def\csname LTa\endcsname{\color{black}}%
      \expandafter\def\csname LT0\endcsname{\color[rgb]{1,0,0}}%
      \expandafter\def\csname LT1\endcsname{\color[rgb]{0,1,0}}%
      \expandafter\def\csname LT2\endcsname{\color[rgb]{0,0,1}}%
      \expandafter\def\csname LT3\endcsname{\color[rgb]{1,0,1}}%
      \expandafter\def\csname LT4\endcsname{\color[rgb]{0,1,1}}%
      \expandafter\def\csname LT5\endcsname{\color[rgb]{1,1,0}}%
      \expandafter\def\csname LT6\endcsname{\color[rgb]{0,0,0}}%
      \expandafter\def\csname LT7\endcsname{\color[rgb]{1,0.3,0}}%
      \expandafter\def\csname LT8\endcsname{\color[rgb]{0.5,0.5,0.5}}%
    \else
      \def\colorrgb#1{\color{black}}%
      \def\colorgray#1{\color[gray]{#1}}%
      \expandafter\def\csname LTw\endcsname{\color{white}}%
      \expandafter\def\csname LTb\endcsname{\color{black}}%
      \expandafter\def\csname LTa\endcsname{\color{black}}%
      \expandafter\def\csname LT0\endcsname{\color{black}}%
      \expandafter\def\csname LT1\endcsname{\color{black}}%
      \expandafter\def\csname LT2\endcsname{\color{black}}%
      \expandafter\def\csname LT3\endcsname{\color{black}}%
      \expandafter\def\csname LT4\endcsname{\color{black}}%
      \expandafter\def\csname LT5\endcsname{\color{black}}%
      \expandafter\def\csname LT6\endcsname{\color{black}}%
      \expandafter\def\csname LT7\endcsname{\color{black}}%
      \expandafter\def\csname LT8\endcsname{\color{black}}%
    \fi
  \fi
  \setlength{\unitlength}{0.0500bp}%
  \begin{picture}(8640.00,4320.00)%
    \gplgaddtomacro\gplbacktext{%
      \csname LTb\endcsname%
      \put(528,691){\makebox(0,0)[r]{\strut{}$10^{-5}$}}%
      \put(528,1364){\makebox(0,0)[r]{\strut{}$10^{-4}$}}%
      \put(528,2037){\makebox(0,0)[r]{\strut{}$10^{-3}$}}%
      \put(528,2710){\makebox(0,0)[r]{\strut{}$10^{-2}$}}%
      \put(528,3383){\makebox(0,0)[r]{\strut{}$10^{-1}$}}%
      \put(528,4056){\makebox(0,0)[r]{\strut{}$10^{0}$}}%
      \put(660,440){\makebox(0,0){\strut{}$10^{-2}$}}%
      \put(1748,440){\makebox(0,0){\strut{}$10^{-1}$}}%
      \put(2836,440){\makebox(0,0){\strut{}$10^{0}$}}%
      \put(3924,440){\makebox(0,0){\strut{}$10^{1}$}}%
      \put(286,2358){\rotatebox{90}{\makebox(0,0){\strut{}$\epsilon$}}}%
      \put(2292,110){\makebox(0,0){\strut{}$\nu=N/V$}}%
    }%
    \gplgaddtomacro\gplfronttext{%
    }%
    \gplgaddtomacro\gplbacktext{%
      \csname LTb\endcsname%
      \put(4848,691){\makebox(0,0)[r]{\strut{}$10^{-5}$}}%
      \put(4848,1364){\makebox(0,0)[r]{\strut{}$10^{-4}$}}%
      \put(4848,2037){\makebox(0,0)[r]{\strut{}$10^{-3}$}}%
      \put(4848,2710){\makebox(0,0)[r]{\strut{}$10^{-2}$}}%
      \put(4848,3383){\makebox(0,0)[r]{\strut{}$10^{-1}$}}%
      \put(4848,4056){\makebox(0,0)[r]{\strut{}$10^{0}$}}%
      \put(4980,440){\makebox(0,0){\strut{}$10^{-2}$}}%
      \put(6068,440){\makebox(0,0){\strut{}$10^{-1}$}}%
      \put(7156,440){\makebox(0,0){\strut{}$10^{0}$}}%
      \put(8244,440){\makebox(0,0){\strut{}$10^{1}$}}%
      \put(4606,2358){\rotatebox{90}{\makebox(0,0){\strut{}$\epsilon$}}}%
      \put(6612,110){\makebox(0,0){\strut{}$\nu=N/V$}}%
    }%
    \gplgaddtomacro\gplfronttext{%
      \csname LTb\endcsname%
      \put(6251,1724){\makebox(0,0)[r]{\strut{}$\sigma^{2} = 0.1$}}%
      \csname LTb\endcsname%
      \put(6251,1504){\makebox(0,0)[r]{\strut{}$\sigma^{2} = 0.01$}}%
      \csname LTb\endcsname%
      \put(6251,1284){\makebox(0,0)[r]{\strut{}$\sigma^{2} = 0.001$}}%
      \csname LTb\endcsname%
      \put(6251,1064){\makebox(0,0)[r]{\strut{}$\sigma^{2} = 0.0001$}}%
    }%
    \gplgaddtomacro\gplbacktext{%
      \csname LTb\endcsname%
      \put(674,840){\makebox(0,0)[r]{\strut{}}}%
      \put(674,1172){\makebox(0,0)[r]{\strut{}}}%
      \put(674,1505){\makebox(0,0)[r]{\strut{}}}%
      \put(674,1837){\makebox(0,0)[r]{\strut{}}}%
      \put(2564,605){\makebox(0,0){\strut{}}}%
    }%
    \gplgaddtomacro\gplfronttext{%
    }%
    \gplbacktext
    \put(0,0){\includegraphics{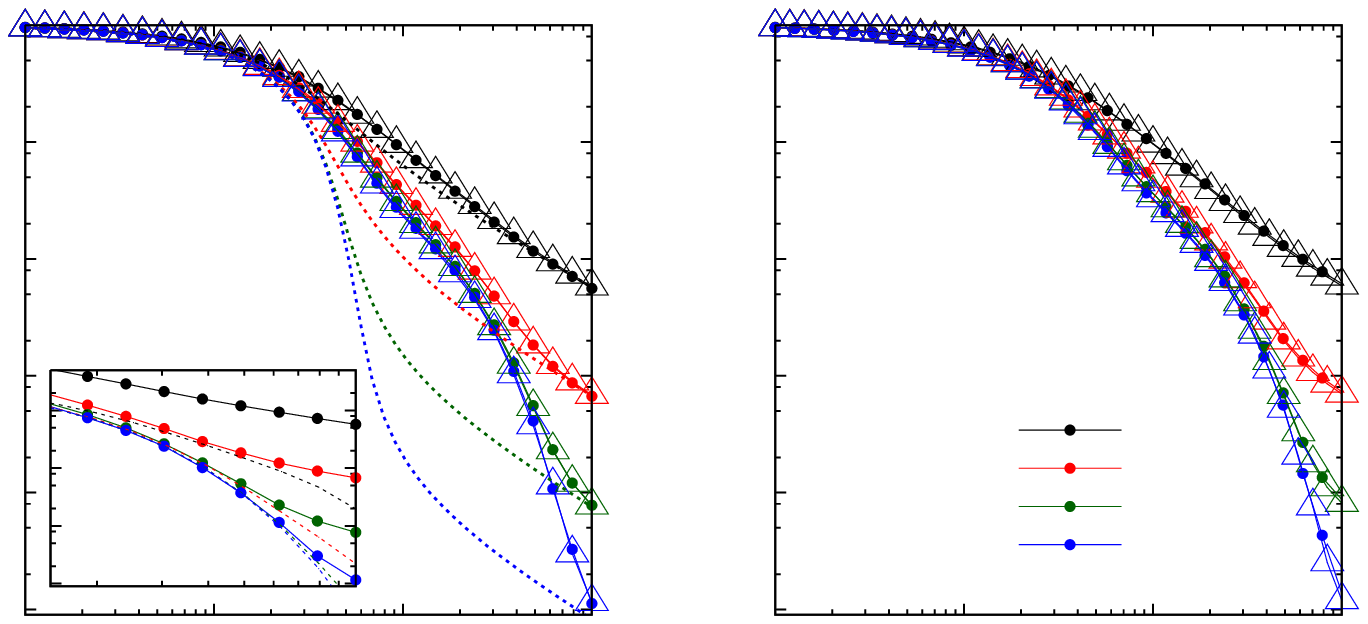}}%
    \gplfronttext
  \end{picture}%
\endgroup

%% file: replica.bbl
\begin{thebibliography}{10}

\bibitem{Seung1992}
H.~S. Seung, H.~Sompolinsky, and N.~Tishby.
\newblock Statistical mechanics of learning from examples.
\newblock {\em Phys Rev A}, 45(8):6056--6091, Apr 1992.

\bibitem{Amari1992}
S.~Amari, N.~Fujita, and S.~Shinomoto.
\newblock Four types of learning curves.
\newblock {\em Neural Comput.}, 4(4):605--618, Jul 1992.

\bibitem{Watkin1993}
T.~L.~H. Watkin, A.~Rau, and M.~Biehl.
\newblock The statistical mechanics of learning a rule.
\newblock {\em Rev. Mod. Phys.}, 65(2):499--556, Apr 1993.

\bibitem{Opper1995}
M.~Opper and D.~Haussler.
\newblock Bounds for predictive errors in the statistical mechanics of
  supervised learning.
\newblock {\em Phys. Rev. Lett.}, 75(20):3772--3775, Nov 1995.

\bibitem{Haussler1996}
D.~Haussler, M.~Kearns, H.~S. Seung, and N.~Tishby.
\newblock Rigorous learning curve bounds from statistical mechanics.
\newblock {\em Mach. Learn.}, 25(2--3):195--236, Nov 1996.

\bibitem{Freeman1997}
J.~Freeman and D.~Saad.
\newblock Dynamics of on-line learning in radial basis networks.
\newblock {\em Phys. Rev. E}, 56(1):907--918, Jul 1997.

\bibitem{Rasmussen2005}
C.~E. Rasmussen and C.~K.~I. Williams.
\newblock {\em Gaussian Processes for Machine Learning}.
\newblock MIT Press, Nov 2005.

\bibitem{Sollich1999a}
P.~Sollich.
\newblock Learning curves for {G}aussian processes.
\newblock In M.~S. Kearns, S.~A. Solla, and D~A Cohn, editors, {\em Advances in
  Neural Information Processing Systems}, volume~11, pages 344--350, 1999.

\bibitem{Sollich1999b}
P.~Sollich.
\newblock Approximate learning curves for {G}aussian processes.
\newblock In {\em International Conference on {A}rtificial {N}eural
  {N}etworks}, volume~9, pages 437--442, 1999.

\bibitem{Opper1999}
M.~Opper and F.~Vivarelli.
\newblock General bounds on {B}ayes errors for regression with {G}aussian
  processes.
\newblock In M.~S. Kearns, S.~A. Solla, and D~A Cohn, editors, {\em Advances in
  Neural Information Processing Systems}, volume~11, pages 302--308, 1999.

\bibitem{Williams2000}
C.~Williams and F.~Vivarelli.
\newblock Upper and lower bounds on the learning curve for {G}aussian
  processes.
\newblock {\em Mach. Learn.}, 40:77--102, Jul 2000.

\bibitem{Malzahn2003}
D.~Malzahn and M.~Opper.
\newblock Learning curves and bootstrap estimates for inference with {G}aussian
  processes: A statistical mechanics study.
\newblock {\em Complexity}, 8(4):57--63, Mar 2003.

\bibitem{Sollich2002a}
P.~Sollich.
\newblock Gaussian process regression with mismatched models.
\newblock In S~Becker T~G~Dietterich and Z~Ghahramani, editors, {\em Advances
  in Neural Information Processing Systems}, volume~14, pages 519--526, 2002.

\bibitem{Sollich2002b}
P.~Sollich and A.~Halees.
\newblock Learning curves for {G}aussian process regression: Approximations and
  bounds.
\newblock {\em Neural Comput.}, 14(6):1393--1428, Jun 2002.

\bibitem{Sollich2005}
P.~Sollich and C.~K.~I. Williams.
\newblock Using the equivalent kernel to understand {G}aussian process
  regression.
\newblock In L.~K. Saul, Y.~Weiss, and L.~Bottou, editors, {\em Advances in
  Neural Information Processing Systems}, volume~17, pages 1313--1320, 2005.

\bibitem{Kondor2002}
R.~Kondor and J.~Lafferty.
\newblock Diffusion kernels on graphs and other discrete structures.
\newblock In C.~Sammut and Hoffmann~A. G., editors, {\em Proceedings of the
  19th International Conference on Machine Learning (ICML)}, pages 315--322,
  Jul 2002.

\bibitem{Smola2003}
A.~Smola and R.~Kondor.
\newblock Kernels and regularization on graphs.
\newblock In B~Sch\"olkopf and M.~K. Warmuth, editors, {\em Lect. Notes Artif.
  Int.}, volume 2777, pages 144--158, 2003.

\bibitem{Chung1996}
F.~K. Chung.
\newblock {\em Spectral graph theory}, volume~92 of {\em Regional conference
  series in mathematics}.
\newblock American Mathematical Society, Dec 1996.

\bibitem{Sollich2009}
P.~Sollich, M.~J. Urry, and C.~Coti.
\newblock Kernels and learning curves for {G}aussian process regression on
  random graphs.
\newblock In Y.~Bengio, D.~Schuurmans, J.~Lafferty, C.~K.~I. Williams, and
  A.~Culotta, editors, {\em Advances in Neural Information Processing Systems},
  volume~22, pages 1723--1731, 2009.

\bibitem{Min2009}
R.~Min, R.~Kuang, A.~Bonner, and Z.~Zhang.
\newblock Learning random--walk kernels for protein remote homology
  identification and motif discovery.
\newblock In H.~Park, S.~Parthasarathy, H.~Liu, and Z.~Obradovic, editors, {\em
  SIAM Proc. S.}, 2009.

\bibitem{Lippert2010}
C.~Lippert, Z.~Ghahramani, and K.~M. Borgwardt.
\newblock Gene function prediction from synthetic lethality networks via
  ranking on demand.
\newblock {\em Bioinformatics}, 26(7):912--918, Apr 2010.

\bibitem{Gao2009}
C.~L. Gao, X.~Dang, Y.~X. Chen, and D.~Wilkins.
\newblock Graph ranking for exploratory gene data analysis.
\newblock {\em BMC Bioinformatics}, 10:S19, 2009.

\bibitem{Malzahn2005}
D.~Malzahn and M.~Opper.
\newblock A statistical physics approach for the analysis of machine learning
  algorithms on real data.
\newblock {\em J. Stat. Mech. Theor. Exp.}, page P11001, Nov 2005.

\bibitem{Opper2002}
M.~Opper and D.~Malzahn.
\newblock A variational approach to learning curves.
\newblock In T.~G. Dietterich, S.~Becker, and Z.~Ghahramani, editors, {\em
  Advances in Neural Information Processing Systems}, volume~14, pages
  463--469, 2002.

\bibitem{Kuhn2007}
R.~K{\"u}hn, J.~{v}an Mourik, M.~Weigt, and A.~Zippelius.
\newblock Finitely coordinated models for low-temperature phases of amorphous
  systems.
\newblock {\em J. Phys. A}, 40(31):9227--9252, Aug 2007.

\bibitem{UrryTBA}
M.~J. Urry and P.~Sollich.
\newblock Random walk kernels and matched learning curves for {G}aussian
  process regression on random graphs.
\newblock To be published.

\bibitem{Urry2010}
M.~J. Urry and P.~Sollich.
\newblock Exact learning curves for {G}aussian process regression on large
  random graphs.
\newblock In J.~Lafferty, C.~K.~I. Williams, J.~Shawe-Taylor, R.S. Zemel, and
  A.~Culotta, editors, {\em Advances in Neural Information Processing Systems},
  volume~23, pages 2316--2324, 2010.

\bibitem{Mezard2001}
M.~Mezard and G.~Parisi.
\newblock The {B}ethe lattice spin glass revisited.
\newblock {\em Eur. Phys. J. B}, 20(2):217--233, Mar 2001.

\bibitem{Mezard1987}
M.~M\'ezard, G.~Parisi, and M.~Virasoro.
\newblock {\em Spin Glass Theory and Beyond}.
\newblock World Scientific, 1987.

\bibitem{FontClosTBA}
F.~Font-Clos, F.~A. Massucci, and I.~P. Castillo.
\newblock A weighted belief-propagation algorithm to estimate volume-related
  properties of random polytopes.
\newblock arXiv:1208.1295v1.

\bibitem{Erdos1959}
P.~Erd{\H{o}}s and A.~R{\'{e}}nyi.
\newblock On random graphs, {I}.
\newblock {\em Publicationes Mathematicae (Debrecen)}, 6:290--297, 1959.

\bibitem{Britton2006}
T.~Britton, M.~Deijfen, and A.~Martin-Loeff.
\newblock Generating simple random graphs with prescribed degree distribution.
\newblock {\em Journal of Statistical Physics}, 124(6):1377--1397, September
  2006.

\bibitem{Rogers2010}
T.~Rogers, C.~Vicente, K.~Takeda, and I.~Castillo.
\newblock Spectral density of random graphs with topological constraints.
\newblock {\em J. Phys. A}, 43(19):195002, May 2010.

\bibitem{Kuhn2011}
R.~K\"{u}hn and J.~{v}an Mourik.
\newblock Spectra of modular and small-world matrices.
\newblock {\em J. Phys. A}, 44(16):165205, Apr 2011.

\bibitem{Hager1989}
W.W. Hager.
\newblock Updating the inverse of a matrix.
\newblock {\em SIAM Rev.}, 31(2):221--239, Jun 1989.

\end{thebibliography}
